\documentclass[lettersize,journal]{IEEEtran}
\usepackage{amsmath,amsfonts}
\usepackage{algorithmic}
\usepackage{algorithm}
\usepackage{array}
\usepackage[caption=false,font=normalsize,labelfont=sf,textfont=sf]{subfig}
\usepackage{textcomp}
\usepackage{stfloats}
\usepackage{url}
\usepackage{verbatim}
\usepackage{graphicx}
\usepackage{cite}
\usepackage{mlmath}
\begin{document}


\title{Complexity Control Facilitates Reasoning-Based Compositional Generalization in Transformers}


\author{Zhongwang Zhang, Pengxiao Lin, Zhiwei Wang, Yaoyu Zhang, Zhi-Qin John Xu 
\IEEEcompsocitemizethanks{\IEEEcompsocthanksitem The authors are with the Institute of Natural Sciences, School of Mathematical Sciences,  MOE-LSC, Shanghai Jiao Tong University, Shanghai, 200240, China. 
Y. Zhang is also with the School of Artificial Intelligence, Shanghai Jiao Tong University, Shanghai, 200240, China.
Z.-Q. J. Xu is also with the School of Artificial Intelligence, Shanghai Jiao Tong University, Center for LLM, Institute for Advanced Algorithms Research, Shanghai Seres Information Technology Co., Ltd, Shanghai 200040, China.
\IEEEcompsocthanksitem Zhi-Qin John Xu is the corresponding author (E-mail: xuzhiqin@sjtu.edu.cn).}}



\maketitle

\begin{abstract}
Transformers have demonstrated impressive capabilities across various tasks, yet their performance on compositional problems remains a subject of debate. In this study, we investigate the internal mechanisms underlying Transformers’ behavior in compositional tasks. We find that complexity control strategies—particularly the choice of parameter initialization scale and weight decay—significantly influence whether the model learns primitive-level rules that generalize out-of-distribution (reasoning-based solutions) or relies solely on memorized mappings (memory-based solutions). By applying masking strategies to the model's information circuits and employing multiple complexity metrics, we reveal distinct internal working mechanisms associated with different solution types. Further analysis reveals that reasoning-based solutions exhibit a lower complexity bias, which aligns with the well-studied neuron condensation phenomenon. This lower complexity bias is hypothesized to be the key factor enabling these solutions to learn reasoning rules. We validate these conclusions across multiple real-world datasets, including image generation and natural language processing tasks, confirming the broad applicability of our findings.
\end{abstract}


\section{Introduction}

Large-scale transformers~\cite{vaswani2017attention} demonstrate unprecedented capabilities~\cite{achiam2023gpt,brown2020language,choi2021chatgpt,teubner2023welcome}, even noted as ``sparks of AGI"~\cite{bubeck2023sparks}. Despite their success, significant gaps remain in their reasoning abilities, particularly when tasked with systematically handling novel scenarios. For instance, language models often struggle to consistently apply logical rules or extend their knowledge to unfamiliar contexts, highlighting their limitations in systematic reasoning, a critical aspect of human cognition~\cite{marcus2003algebraic, smolensky2022neurocompositional}.

A key challenge within systematic reasoning is compositional reasoning, \textit{i.e.}, the ability to generalize from known concepts to novel combinations. Humans demonstrate remarkable proficiency in this domain, and once a child learns the concept of ``skip'', they can easily understand extensions like ``skip backwards'' or ``skip around a cone twice''~\cite{fodor1988connectionism}. This compositional skill, integral to human reasoning, has been a central point of contention for neural networks. While modern models achieve remarkable performance on various tasks, their ability to perform compositional reasoning, especially in out-of-distribution (OOD) scenarios, remains limited~\cite{keysers2019measuring,yu2020assessing,hupkes2020compositionality,press2023measuring,kim2020cogs,wang2024grokked}. 

This raises critical open questions about how to faithfully interpret transformers' capabilities on compositional tasks: do transformers genuinely learn compositional primitives within the data, or do they primarily rely on memorizing input-output mappings? When they fail on compositional tasks, are their errors systematic, or do they reflect a deeper lack of reasoning structure? Addressing these questions is crucial for understanding their mechanisms and limitations. Furthermore, understanding these issues can inform strategies to enhance model generalization, particularly in OOD scenarios.

\begin{figure}[h]
\centering
\includegraphics[width=0.94\linewidth]{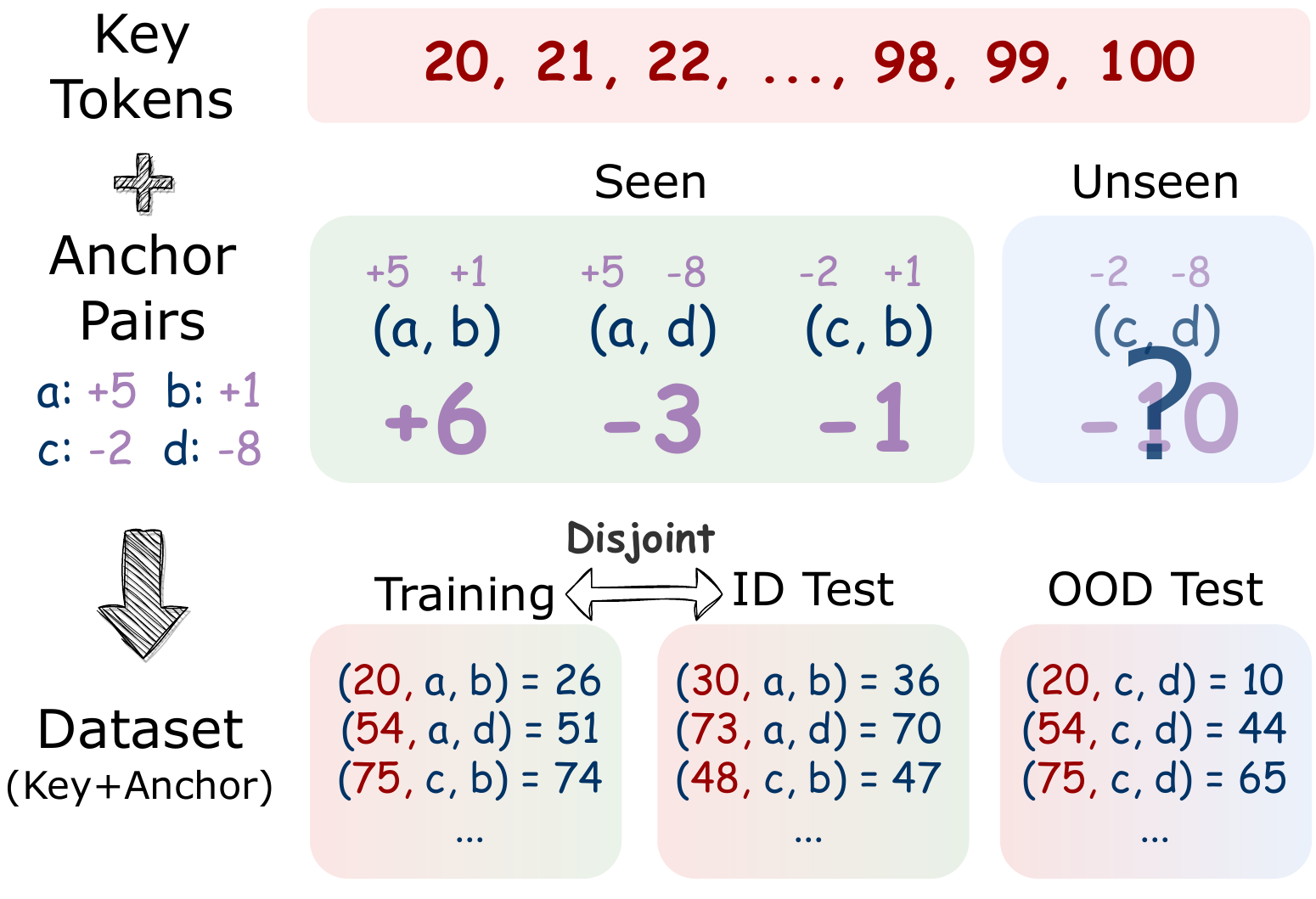}
\caption{The schematic diagram of dataset design. The dataset consists of key tokens, integers ranging from 20 to 100, combined with anchor pairs, which represent arithmetic operations. Each data point is formed by pairing a key token with an anchor pair, and the target is the result of applying the anchor pair's operations to the key token. The training set and ID test set uses disjoint combinations of seen anchor pairs, while the OOD test set involves unseen anchor pairs formed from operations seen during training. The model is trained on the training set and evaluated on both ID and OOD test sets to assess generalization.
}
\label{fig:intro}
\end{figure}

To investigate the limitations and potentials of transformers in compositional reasoning, we adopt a synthetic experimental approach inspired by model-experimental systems in natural sciences. Analogous to how simplified model organisms in biology, like fruit flies or mice, enable mechanistic insights into complex neural systems, our setup leverages anchor functions as interpretable and controllable benchmarks to study reasoning in transformers. A key feature of our dataset design is its precise partitioning, which ensures a perfect separation among training data, in-distribution (ID) test data, and OOD test data as illustrated schematically in Fig.~\ref{fig:intro}. A detailed description of the data construction process is provided in Fig.~\ref{fig:data_intro}. This setup allows us to systematically probe models' generalization capabilities in distinct scenarios, revealing not only the conditions under which compositional generalization succeeds but also the mechanistic underpinnings of their reasoning behavior. Furthermore, this approach ensures relevance to real-world challenges, where reasoning over unseen combinations of known elements is often critical, such as in language understanding, decision-making, and generative tasks. Our main contributions follow.\footnote{Code is available at: \url{https://github.com/sjtuzzw/complexity_control}.}

\textbf{Data Construction for Compositional Reasoning (Section~\ref{sec:def}):} We propose a novel data generation framework that carefully structures input sequences to clearly distinguish between training data, ID test data, and OOD test data. This enables a rigorous examination of model generalization capabilities across different scenarios, from seen to unseen combinations of operations, offering a clean separation that supports the investigation of compositional generalization.

\textbf{Impact of Complexity Control on Model Reasoning (Section~\ref{sec:phase}):} In this study, complexity control refers to the strategic adjustment of parameter initialization scales and weight decay coefficients to influence the model's complexity and reasoning abilities. Our experiments reveal that strong complexity control, namely small initialization and large weight decay coefficients, significantly enhances the model's ability to learn compositional primitives in complex tasks. This improvement is particularly evident in OOD generalization, where the model is better able to handle unseen combinations of operations. We categorize the results into three phases, demonstrating how different levels of complexity control lead to varying reasoning abilities and generalization patterns.

\textbf{Mechanism Analysis and Complexity Validation through Masking and Structural Probing (Sections~\ref{sec:mechanism} and \ref{sec:complexity}):} We conduct a comprehensive analysis of the internal mechanisms of transformer models by applying masking strategies to various components, including key tokens and anchor pairs, to elucidate how different phases of the model fit the data. This investigation reveals the underlying factors driving the differences in reasoning capabilities across phases. Furthermore, we assess model complexity through neuron condensation, namely weights of different neurons condensing in a few directions, structural organization of embedding matrix, and rank-based analysis. These approaches shed light on how model complexity influences reasoning behavior and validate the role of complexity control in guiding models toward simpler, rule-based representations.

\textbf{Real-World Application and Model Behavior Prediction (Sections~\ref{sec:real_task} and \ref{sec:pred}):} Finally, we demonstrate the practical relevance of our findings by applying our approach to several real-world tasks, including diffusion models and language processing tasks. With complexity control, we observe notable improvements in OOD generalization performance on these tasks. Furthermore, based on our understanding of the model's behavior across different settings, we are able to predict its behavior during the training process.

This study is an extension of our conference paper~\cite{zhang2024initialization}. The main improvements over the preliminary version are: \textit{i)} We have extended the experimental setup and conducted new experiments, resulting in more refined phases compared to the original study. Additionally, by introducing a novel masking strategy, we can clearly delineate the differences in model mechanisms across these phases. These enhancements enable a more detailed analysis of model behavior and provide a clearer understanding of both ID and OOD generalization. \textit{ii)} Through a unified complexity analysis, we combine multiple metrics and employ the rank-based perspective to gain deeper insights into how complexity control affects the model. This complexity analysis is closely linked to the model's ID and OOD performance and the underlying mechanisms across different phases. With these insights, we can predict the training behaviors of models under various configurations. \textit{iii)} Additionally, we have extended the validation to image generation and real natural language tasks, thereby enhancing the practical relevance and generality of our findings compared to the original paper's focus on synthetic language data.

\section{Related Work}

Recent advancements in large language models (LLMs) have showcased remarkable capabilities, often surpassing human performance~\cite{fu2022does, wei2022emergent}. However, despite their impressive performance on single-step reasoning tasks~\cite{srivastava2022beyond}, transformers struggle with multi-step compositional tasks and OOD generalization~\cite{csordas2021neural, dziri2024faith,   hupkes2018learning,lepori2023break, okawa2023compositional, yun2022vision,wang2024towards, csordas2022ctl}. Ramesh et al.~\cite{ramesh2023capable} show that transformers trained to directly compose capabilities struggle to generalize to OOD tasks with a synthetic task. Liu et al.~\cite{liu2022transformers} suggest that shallow transformers learn shortcuts during training, leading to poor OOD generalization. Numerous studies have explored various approaches to address these limitations, such as encouraging explicit reasoning step generation within a single generation~\cite{wei2022chain}, leveraging LLMs to generate reasoning steps iteratively~\cite{creswell2022selection, creswell2022faithful}. Despite these efforts, achieving complete mastery of compositional tasks remains a significant challenge for vanilla transformers. A series of works study the internal mechanisms of language models and improve the capabilities of language models~\cite{wang2024improving,wang2024understanding,wang2023label,cao2024graphinsight}. In order to clearly study the behaviors and internal mechanisms of language models, 
Zhang et al.~\cite{zhang2024anchor} introduced anchor functions as benchmark functions for investigating transformer behavior. Our work builds on the anchor function setting to explore how different initialization scales affect model solutions and mechanisms.

Compositional tasks are equally crucial in diffusion models. Due to the difficulty of including all possible combinations of concepts in a dataset, the success of compositional tasks determines whether diffusion models can be applied to complex real-world scenarios for image generation. A series of works \cite{marcus2022very,leivada2023dall,conwell2022testing,gokhale2022benchmarking,du2023reduce,liu2022compositional,fengtraining} have investigated the compositional generalization capabilities of off-the-shelf text-conditioned diffusion models. Modern diffusion models are often capable of composing complex concepts to generate entirely novel objects, but they also fail unpredictably when combining seemingly equally complex concepts. A line of research~\cite{okawa2024compositional,liang2024diffusion,yang2024dynamics,parkemergence} has specifically examined compositional generalization in diffusion models through synthetic experimental setups. Okawa et al.~\cite{okawa2024compositional} introduce a ``concept graph'' framework to analyze how models learn to combine basic attributes, such as shape and color, to generate OOD samples. They emphasize an interesting failure mode in specific settings, where the model loses its ability to generalize compositionally, a limitation that persists even after fine-tuning. In our work, we demonstrate that adjusting initialization scales effectively resolves this failure mode. This result highlights the universal applicability of complexity control across different models and tasks, providing new insights into improving the compositionality of generative models.

The parameter initialization of the network is important to determine the fitting result of the network~\cite{arora2019exact,chizat_global_2018,zhang_type_2019,e2020comparative,jacot_neural_2018,mei_mean_2018,rotskoff_parameters_2018,sirignano_mean_2020,williams_gradient_2019}. Luo et al.~\cite{luo2021phase}, Zhou et al.~\cite{zhou2022empirical} mainly identify the linear regime and the condensed regime for two-layer and three-layer wide ReLU NNs, respectively. A series of works suggests that the condensed networks are often accompanied by good generalization ability of the model~\cite{zhang2022linear,zhang2023loss,zhang2023stochastic,zhang2024implicit}. A line of works links the grokking phenomenon with the improvement of generalization capability~\cite{power2022grokking,wang2024grokked,gopalani2024transformers} and points out that the initialization scale has an important influence on the occurrence of grokking~\cite{liu2022omnigrok}. 
Most of these studies focus on ID generalization, investigating how models transfer to new data within the same task. Some suggest that only larger initialization scales impair generalization compared to default initialization. Our work focuses on OOD generalization, which demands stronger reasoning capabilities to uncover underlying patterns. While effective ID generalization is achievable, we find significant variation in models' reasoning abilities and mechanisms under different configurations, often requiring initialization scales much smaller than the default to achieve better OOD generalization.
Recent studies also investigate the impact of initialization on the training process of LLMs~\cite{huang2020improving,liu2020understanding,trockman2023mimetic,wang2024deepnet,zhang2019improving,zhu2021gradinit}. These works primarily focus on how the initialization scale affects the stability of the training process and plays a crucial role in ensuring smooth and effective training of LLMs. In our work, we find that different initialization scales can significantly influence a model’s capacity to both memorize and reason about compositional tasks, highlighting the profound impact of initialization on the final performance and underlying mechanisms of the trained models.

\section{Definitions}\label{sec:def}

\begin{figure}[h]
\centering
\includegraphics[width=0.99\linewidth]{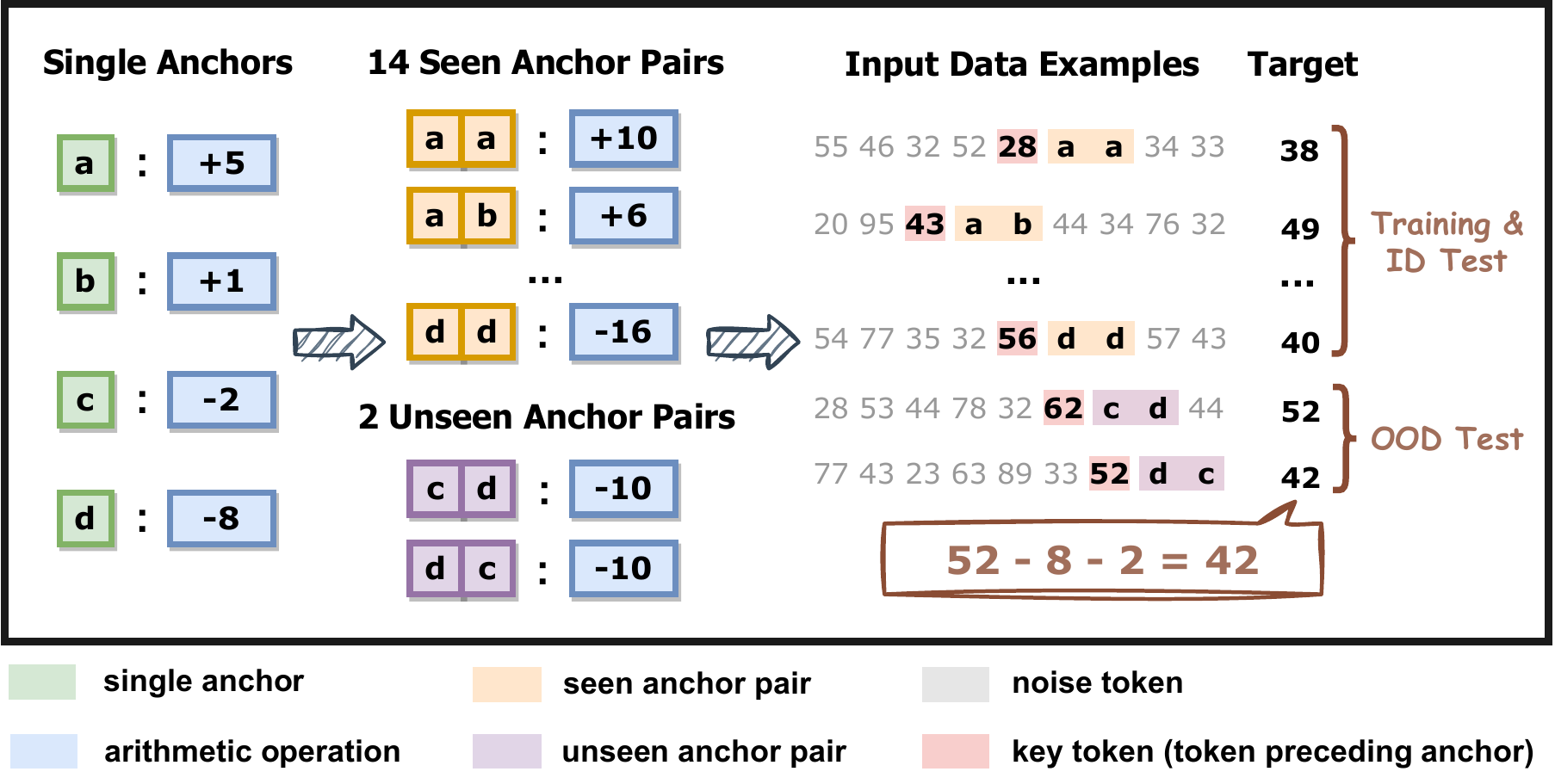}
\caption{Experimental setup for the compositional task. Left: The single anchors (\textit{i.e.}, $a$, $b$, $c$, $d$) correspond to specific arithmetic operations. Middle: During training, 14 out of the 16 possible anchor pairs are seen in the training set, and the remaining pairs ($c$, $d$), ($d$, $c$) are held out as unseen tasks (does not appear during training). Right: The input sequences comprise an anchor pair, a key token preceding the anchor pair, and noise tokens unrelated to the target. We construct mutually exclusive training and ID test sets using data generated from 14 seen anchor pairs (the specific partitioning method is detailed in Appendix~\ref{data_split}), while data from the remaining 2 unseen anchor pairs is used to form the OOD test set.
}
\label{fig:data_intro}
\end{figure}

We introduce a set of key definitions that will be used throughout the paper. Fig.~\ref{fig:data_intro} provides a brief description of our definitions.

\subsection{Two-anchor composite function} \label{sec:comp_func}

The two-anchor composite function applies two sequential operations to a specific token. This setup allows us to dissect whether the model learns these operations as discrete, composable rules or merely memorizes specific input-output mappings. A two-anchor composite function $f(X): \mathbb{R}^{n} \rightarrow \mathbb{R}$ is defined as
\begin{equation}
f(x_1, \ldots, x_n) = g\left(g(x_{i-1}; x_i); x_{i+1}\right), 
\label{equ:com_anchor}
\end{equation}
where $x_i$, $x_{i+1}\in A$. Here, the input sequence $X = (x_1, \ldots, x_n)$ comprises $n$ tokens. An anchor set $A = \{a_1, a_2, \ldots, a_J\}$ is designated, with each token $a_k \in A$ corresponds to a function $g(x; a_k)$.
In each $X$, one and only one pair of two consecutive elements belong to $A$, such as $x_i, x_{i+1} \in A$. We refer to the token immediately preceding the anchor pair as the key token. To simplify notation, we denote the two-anchor composite function as $f(x_{i-1}; x_{i}, x_{i+1})$ to emphasize the anchor pair $(x_{i}, x_{i+1})$ and key token $x_{i-1}$. 

In this work, we set the anchor set $A = \{a, b, c, d\}$. Each anchor token corresponds to a specific function:
\begin{equation}
\begin{aligned}
g(x;a) &= x+5, \quad g(x;b) = x+1, \\
g(x;c) &= x-2, \quad g(x;d) = x-8.
\end{aligned}
\label{equ:anchor}
\end{equation}

{\bf{Example.}} Suppose we have an input sequence $X=(23,a,b,43,46,74,54,44,72)$. In this sequence, the second and third tokens (\textit{i.e.}, tokens $a$ and $b$) belong to the anchor set $A$, thus forming an anchor pair $(a,b)$. The token 23, immediately preceding the anchor pair, is called the key token.

For this input sequence $X$, the computation process of the two-anchor composite function is as follows:
\begin{align*}
f(X)&=f(23;a,b)=g(g(23;a);b)\\
&=g(23+5;b)=28+1=29.
\end{align*}

\subsection{Data Generation} \label{sec:dataset}

\textbf{Input data.} In this work, we construct the input dataset using four anchors (\textit{i.e.}, $a$, $b$, $c$, $d$) and numerical tokens sampled from 20 to 100. For each sequence, two anchors are selected (allowing repetition) to form an anchor pair, and the sequence is composed of this anchor pair along with other randomly sampled numerical tokens. The token immediately preceding the anchor pair is designated as the key token, while the remaining tokens are treated as noise tokens unrelated to the target. Noise tokens are primarily used to separate the training set from the ID test dataset. For details, please refer to Appendix~\ref{data_split}. The four anchors form 16 anchor pairs, of which $(c, d)$ and $(d, c)$ are designated as unseen anchor pairs. The training dataset is constructed based on the remaining 14 seen anchor pairs.

\textbf{Target.} The target is the output of the key token processed by the two-anchor composite function, \textit{i.e.}, the corresponding output of the two-anchor composite function. 

\textbf{Dataset Partition.} We divided the dataset into three categories: the training set, the ID test set, and the OOD test set. For the training and ID test sets, the anchor pairs are selected from 14 seen anchor pairs, excluding $(c, d)$ and $(d, c)$. The division is based on combinations of anchor pairs and key tokens; the specific method is detailed in Appendix~\ref{data_split}. For the OOD test set, the anchor pairs are taken from $(c, d)$ or $(d, c)$.

\subsection{Generalization}

The division of the dataset naturally leads to the following two concepts of generalization:

\textbf{ID generalization.} Generalization on the ID test set, where all anchor pairs are seen in the training set.

\textbf{OOD generalization.} Generalization on the OOD test set, where anchor pairs do not appear in the training set. 

\subsection{Model Architecture and Basic Experimental Setups}

For foundational analyses, we employ a single-head transformer model to simplify mechanism explorations. Subsequent experiments extend these findings to the multi-head architecture of GPT-2, ensuring the generality of our conclusions. The following sections will only introduce the architecture of the single-head attention model, as the multi-head case is a natural extension of the single-head model.

The input sequence is represented as a one-hot vector $X^{\mathrm{in}}$. The word embedding $X^{\mathrm{em}}$ and the input to the first transformer block $X^{(1)}$ is calculated as:
\begin{equation}
    X^{\mathrm{em}} = X^{\mathrm{in}}W^{\mathrm{em}} , X^{(1)} = X^{\mathrm{em}} + X^{\mathrm{pos}},
\end{equation}
where $X^{\mathrm{pos}}$ is a trainable positional vector. For the $l$-th layer, the $Q, K, V$ are defined as:
\begin{equation}
    Q^{(l)} = X^{(l)}W^{Q(l)}, K^{(l)} = X^{(l)}W^{K(l)},  V^{(l)} = X^{(l)}W^{V(l)}.
\end{equation}
The attention matrix $\mathrm{Attn}^{(l)}$ and its subsequent output  $X^{\mathrm{qkv}(l)}$ for the $l$-th layer is computed as:
\begin{equation}
\begin{aligned}
    \mathrm{Attn}^{(l)} &= \mathrm{softmax}\left(\frac{Q^{(l)}K^{(l)T}}{\sqrt{d_k}}\right) \text{ (with mask)}, \\
    X^{\mathrm{qkv}(l)} &= \mathrm{Attn}^{(l)} V^{(l)}.
\end{aligned}
\end{equation}

The output of the $l$-th attention layer is obtained as:
\begin{equation}
\begin{aligned}
    X^{\mathrm{ao}(l)} &= \text{LN}(X^{(l)} + X^{\mathrm{qkv}(l)}W^{\mathrm{attn}, l}), \\
    X^{(l+1)} &:= X^{\mathrm{do}(l)} = \text{LN}(\text{MLP}(X^{\mathrm{ao}(l)}) + X^{\mathrm{ao}(l)}).
\end{aligned}
\end{equation}

where ``LN'' refers to Layer Normalization. The final output is obtained by projecting the output of the last layer $X^{\mathrm{do}(L)}$ using a linear projection layer, followed by a softmax operation and argmax to obtain the predicted token.

We use \( f_{\vtheta}(x; a_1, a_2; \vn) \) to denote the output of the neural network for a sequence with \( x \) as the key token, \( (a_1, a_2) \) as the anchor pair, and \( \vn \) as the noise token sequence. At this point, the input sequence is uniquely determined with the values and positional information of the noise tokens in \( \vn \). 

For the basic experimental setups, we use cross-entropy loss on the last token of the sequence and optimize the model using Adam with weight decay. The specific hyperparameters and training details are provided in Appendix \ref{app:setup}.

\subsection{Initialization and Regularization Parameters}

\textbf{Initialization Rate.} The initialization rate, denoted by \(\gamma\), governs the scale of parameter initialization within the model. Specifically, model parameters are initialized by sampling from a normal distribution \(\mathcal{N}\left(0, \left(\frac{1}{d_{\mathrm{in}}^{\gamma}}\right)^2\right)\), where \(d_{\mathrm{in}}\) represents the input dimension. A higher \(\gamma\) results in smaller initialization scales.

\textbf{Weight Decay Coefficient.} The weight decay coefficient is a hyperparameter for \(L_2\) regularization during training. It controls the regularization strength by adding \(\lambda \|W\|_2^2\) to the loss function, where \(W\) are the model's weights.

\section{Phases of Solutions for Composite Functions}\label{sec:phase}

In this section, we explore the learning dynamics of a Transformer-based model, specifically GPT-2~\cite{radford2019language}, on compositional tasks using both ID and OOD datasets as defined in Section~\ref{sec:dataset}. Our investigation focuses on understanding how variations in initialization scales and weight decay coefficients influence the model's generalization capabilities.

Initially, we assess the impact of different initialization scales on the model’s ID and OOD performance, maintaining the weight decay coefficient at a fixed value of 0.01, as depicted in Fig.~\ref{fig:phases}A. The abscissa represents the initialization rate $\gamma$ and the ordinate indicates the accuracy achieved on ID data (blue) and OOD data (red). Based on ID and OOD generalization capabilities, we categorize the different initialization settings into three phases:

\begin{figure}
\centering
\includegraphics[width=1\linewidth]{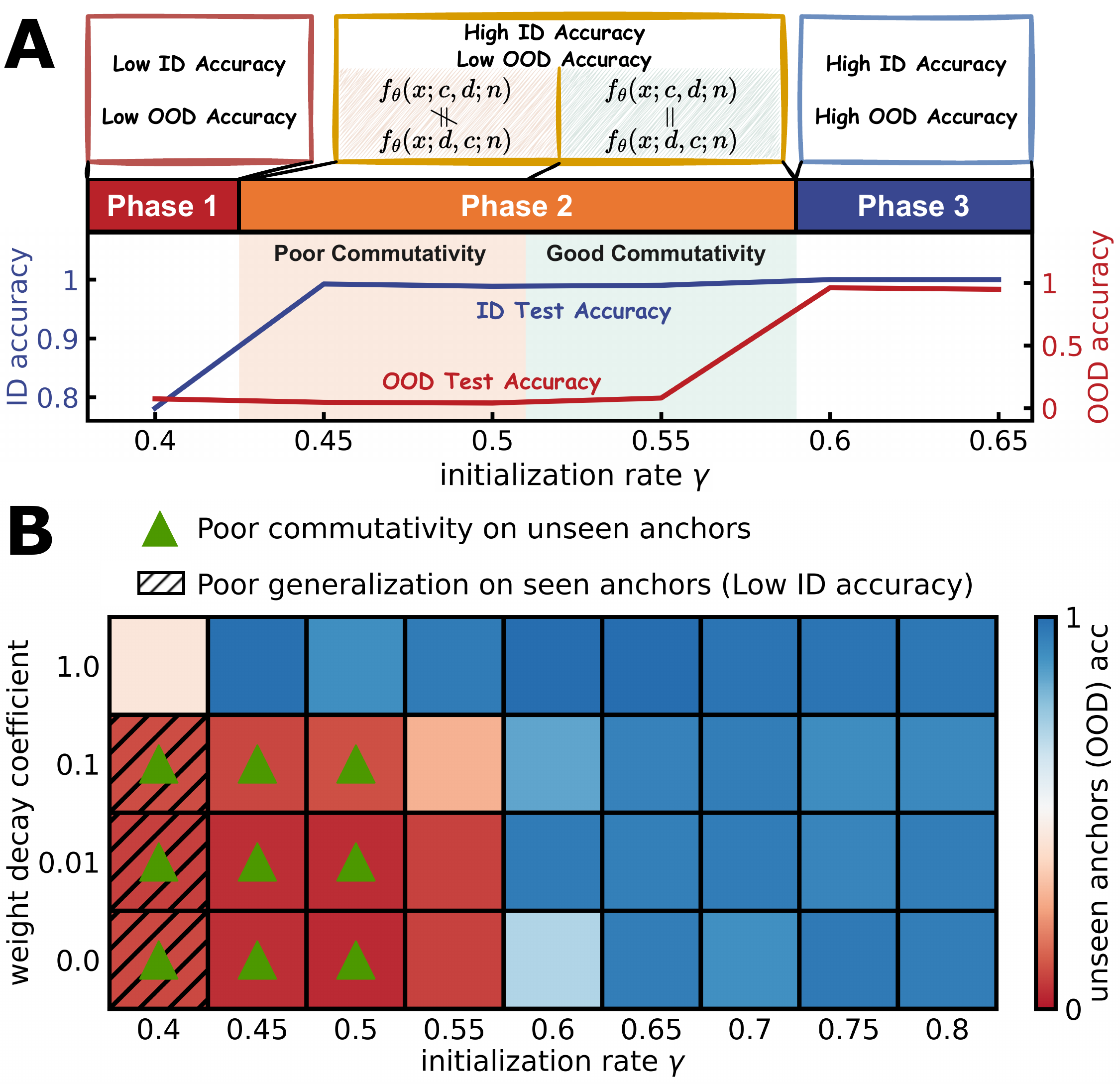}
\caption{\textbf{(A)} ID and OOD generalization of the GPT-2 model on compositional tasks with a fixed weight decay coefficient of 0.01. The abscissa represents the initialization rate $\gamma$, corresponding to the standard deviation $\left(1/{d_{\mathrm{in}}}\right)^{\gamma}$ of the normal distribution used for parameter initialization. The ordinate denotes the accuracy for ID (blue) and OOD (red) data. The different phases are classified based on their ID and OOD generalization abilities.
\textbf{(B)} Heatmap illustrating the GPT-2 model's OOD generalization on compositional tasks, with accuracy on unseen anchor pairs depicted by color intensity. The abscissa matches that of Fig.~\ref{fig:phases}A, while the ordinate represents the weight decay coefficient. Each setting reflects the average results from three independent trials. Striped regions indicate poor ID generalization ($\text{ID accuracy}<90\%$). Green triangles highlight instances of poor commutativity on unseen anchor pairs ($c$, $d$) and ($d$, $c$) ($\text{commutativity probability}<70\%$) when switching anchor pairs.
}
\label{fig:phases}
\end{figure} 

\textbf{Phase 1} (poor ID and poor OOD generalization): The model primarily relies on memorization, with limited ID and OOD generalization capability.

\textbf{Phase 2} (good ID but poor OOD generalization): The model learns composite anchor mappings, achieving good ID generalization but lacking anchor-level abstraction, leading to poor OOD generalization. 

\textbf{Phase 3} (good ID and good OOD generalization):  The model learns individual anchor mappings, understanding task structure from a higher level, enabling strong OOD generalization on unseen anchor pairs.

To further elucidate the model’s reasoning capabilities, we examine its adherence to commutativity, a fundamental property in addition tasks. Specifically, we evaluate whether the model's output remains consistent when interchanging unseen anchor pairs from ($d$, $c$) to ($c$, $d$) while keeping the key and noise tokens fixed.\footnote{We do not focus on the output’s relationship to the target value. For instances with high OOD accuracy, the model inherently satisfies commutativity as the outputs align with target values.} Within Phase 2, commutativity exhibits variability contingent on the initialization rate. With a large initialization rate $\gamma$ (small parameter initialization scales), the model treats symmetric anchor pairs as unified entities, thereby demonstrating good commutativity. Formally, $f_{\vtheta}(x; c, d; \vn) = f_{\vtheta}(x; d, c; \vn)$ holds for the majority of key tokens $x$ and noise sequences $\vn$. For models with small initialization rate $\gamma$ (large parameter initialization scales), they treat each anchor pair as an independent mapping, leading to poor commutativity.

Beyond initialization scales, we investigate the role of the weight decay coefficient in shaping the model's generalization capabilities, as illustrated in Fig.~\ref{fig:phases}B. This analysis examines the combined effects of varying initialization rates $\gamma$ and weight decay coefficients on OOD performance, as indicated by the heatmap colors. Striped regions signify poor ID generalization ($\text{ID accuracy}<90\%$), while green triangles denote instances where commutativity probabilities fall below 70\% when switching unseen anchor pairs from ($d$, $c$) to ($c$, $d$). 

This pattern held consistently across trials, with results averaged over three random seeds. We conclude that as the initialization rate and weight decay coefficient gradually increase, models undergo a transition from basic memorization to a deeper comprehension of the task structure, enhancing reasoning and mastering underlying rules in complex generalization tasks.

We regard the adjustment of the initialization rate and weight decay coefficient as unified aspects of complexity control, since the values of these parameters significantly influence the model's complexity characteristics. Consequently, models with differing complexity levels exhibit substantially different internal mechanisms. In the subsequent sections, we will further explore the mechanisms and complexity differences of models across various phases.

\begin{figure*}[ht]
\centering
\includegraphics[width=0.98\linewidth]{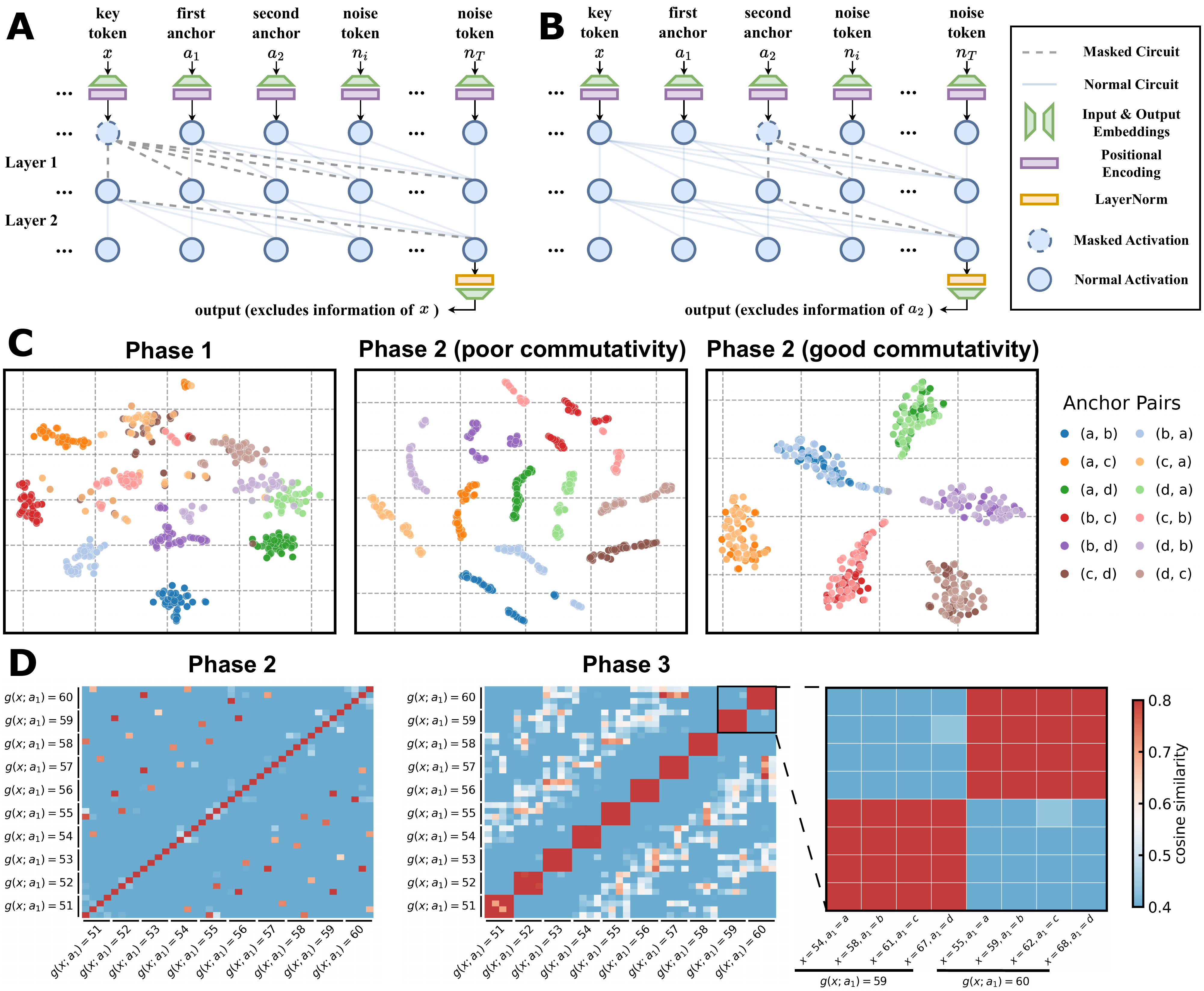}
\caption{\textbf{(A, B)} Masking strategies applied to the transformer model to analyze the contribution of specific tokens. \textbf{(A)} The key token (\( x \)) is masked, removing its information from the model's output, allowing analysis of the output space distributions influenced by anchor pairs (\( a_1, a_2 \)) and noise tokens. \textbf{(B)} The second anchor (\( a_2 \)) is masked to isolate the interaction between the key token (\( x \)) and the first anchor (\( a_1 \)). Dashed gray lines indicate the masked circuits, while solid blue lines represent normal circuits. Hollow nodes depict masked activations, and filled nodes depict normal activations.  
\textbf{(C)} Principal Component Analysis (PCA) applied to the model outputs to visualize the model’s representation of different anchor pairs. For each anchor pair, 50 data points were sampled. Symmetric anchor pairs (\textit{e.g.}, \( (c, d) \) and \( (d, c) \)) are shown in similar colors with different shades to indicate their equivalence. The three phases are achieved by adjusting the initialization rate (\( \gamma \)).  
\textbf{(D)} Cosine similarity matrices between model outputs after masking the second anchor (\( a_2 \)). Each group of four blocks corresponds to inputs with the same \( g(x; a_1) \) value. The right panel provides a magnified view of selected blocks, showing detailed values of \( x \) and \( a_1 \) for corresponding inputs. The color scale represents cosine similarity values. Different phases correspond to different values of the initialization rate \( \gamma \).}
\label{fig:infor_flow}
\end{figure*}

\section{Mechanism Analysis Across Different Phases Through Attention Masking Strategies}\label{sec:mechanism}
In this section, we analyze the mechanisms underlying different phases by selectively masking specific information circuits. This approach allows us to isolate and examine the contributions of key tokens and anchor pairs, systematically investigating the output space and uncovering the model's internal mechanisms across phases. Specifically:
\textit{i)}~We investigate the distribution patterns of different anchor pairs in the output space by masking key token information (Figs.~\ref{fig:infor_flow}A, \ref{fig:infor_flow}C).
\textit{ii)}~We explore the distribution patterns of key tokens with single anchors in the output space by masking the second anchor's information (Figs.~\ref{fig:infor_flow}B, \ref{fig:infor_flow}D).

To achieve this, we mask specific circuits in the model to exclude the information of a particular token (masked activation) from the output, allowing us to study the model's behavior when only other input information is present. Masking is implemented by disabling specific circuits, as indicated by dashed gray lines in Figs.~\ref{fig:infor_flow}A and~\ref{fig:infor_flow}B, while preserving the normal circuits for unmasked tokens (solid blue lines). To preserve the structure of the normal circuits, masking is applied to the attention matrix after the softmax operation. To simplify the analysis, we use a two-layer, single-head transformer model.

\subsection{Mechanism Analysis via Key Token Masking}

Building on the masking strategies described earlier, we analyze the mechanisms underlying different phases of the model by visualizing the output embeddings using TSNE. Fig.~\ref{fig:infor_flow}C shows how the model organizes the representations of key tokens and anchor pairs under varying conditions. For this analysis, we randomly sampled 50 data points for each anchor pair to assess how effectively the model distinguishes between different mappings and whether it adheres to task-specific properties such as commutativity. To enhance clarity, we employed similar colors with varying shades to denote symmetric anchor pairs (\textit{e.g.}, \( (c, d) \) and \( (d, c) \)), indicating their potential equivalence under commutative operations.  

The three phases observed in Fig.~\ref{fig:infor_flow}C are achieved by adjusting the initialization rate (\( \gamma \)), which controls the standard deviation of parameter initialization. In Phase 1 (the left panel of Fig.~\ref{fig:infor_flow}C), the representations of different anchor pairs are poorly organized, with significant overlap between clusters. This indicates that the model fails to capture the distinct mappings associated with each anchor pair, resulting in representational entanglement and poor generalization. The lack of clear structure reflects a reliance on memorization rather than learning the compositional rules underlying the task. As a result, the model exhibits weak performance on both ID and OOD data, demonstrating its inability to generalize even within the training distribution.  

In Phase 2 (the middle and right panels of Fig.~\ref{fig:infor_flow}C), the model begins to capture structural relationships between key tokens and anchor pairs, as evidenced by the emergence of distinct and well-separated clusters. In the middle panel, the model treats symmetric anchor pairs (\textit{e.g.}, \( (c, d) \) and \( (d, c) \)) as independent clusters, failing to recognize their equivalence. This lack of abstraction results in poor commutativity despite improved ID generalization. In contrast, the right panel demonstrates a more advanced model state, where symmetric pairs are successfully merged into single clusters. This structural refinement not only enables good commutativity and enhances generalization but also hints at a transition toward a simpler, rule-based approach for fitting the data. The shift from disorganized to highly structured clusters underscores the model’s tendency to adopt lower complexity and systematic solutions in later phases.

It is worth noting that Phase 3 represents a special case of solutions with good commutativity. Therefore, it is not feasible to analyze the differences between Phase 2 with good commutativity and Phase 3 from the perspective of composite anchors alone. In the next subsection, we focus on analyzing these differences at the level of single anchors to uncover the underlying mechanisms of the model’s task learning.

\subsection{Mechanism Analysis via Second Anchor Masking}

To investigate the mechanisms underlying single-anchor operations and their alignment with the defined single-anchor function \( g(\cdot; \cdot) \) in Equation~(\ref{equ:anchor}), we analyze the distribution patterns of key tokens with single anchors in the output space by masking the second anchor's information, as described in Fig.~\ref{fig:infor_flow}B. In Fig.~\ref{fig:infor_flow}D, the color represents the pairwise cosine similarity between the outputs of the masked model, which isolates the contribution of the key token and the first anchor. Each group of four blocks corresponds to inputs that yield the same value of \( g(x; a_1) \), and the enlarged panel on the right provides detailed input values of \( x \) and \( a_1 \) for these selected blocks.  

In the left panel of Fig.~\ref{fig:infor_flow}D, corresponding to Phase 2, output embeddings with the same value of \( g(x; a_1) \) exhibit low correlation, as reflected by the sparse red diagonal lines and generally low cosine similarity. This observation indicates that the model relies on memorizing composite mappings to fit the data, even when it captures symmetric relationships. In contrast, the middle panel, representing Phase 3, shows significantly higher correlations among inputs with the same \( g(x; a_1) \) value, as indicated by the more prominent red regions in the diagonal blocks. This suggests that the model computes the single-anchor mappings progressively, following the defined function \( g(\cdot; \cdot) \) to produce the final output. The magnified view on the right highlights consistent cosine similarity patterns across input values within each block, further confirming the stepwise reasoning approach in Phase 3.  

These results illustrate a fundamental difference between the two phases: while Phase 2 relies on memorizing composite mappings, Phase 3 demonstrates a reasoning-based strategy by stepwise computing the single-anchor mappings. This shift highlights the model's ability in Phase 3 to generalize OOD data with compositional rules, rather than relying on memorization, and suggests a transition toward simpler and more structured data-fitting strategies.

\section{Model Complexity: A Key Factor in Phase Transitions}\label{sec:complexity}

Building on the analysis of model mechanisms in the previous section, we posit that the model’s complexity preferences play a pivotal role in driving phase transitions. In this section, we examine this phenomenon from two key perspectives: the condensation~\cite{zhou2021towards,zhang2021embedding,zhang2022embedding} of input weights and the structured organization of the word embedding matrix.

Condensation occurs when the input weights of neurons cluster along a few isolated orientations, effectively reducing the model’s complexity and approximating a smaller-scale network. However, parameters like the word embedding matrix, which need to encode distinctions between diverse input tokens, resist condensation and may instead form systematic, rule-based structures. We examine the model's complexity across different phases through these two aspects.

\begin{figure*}[ht]
\centering
\includegraphics[width=0.99\linewidth]{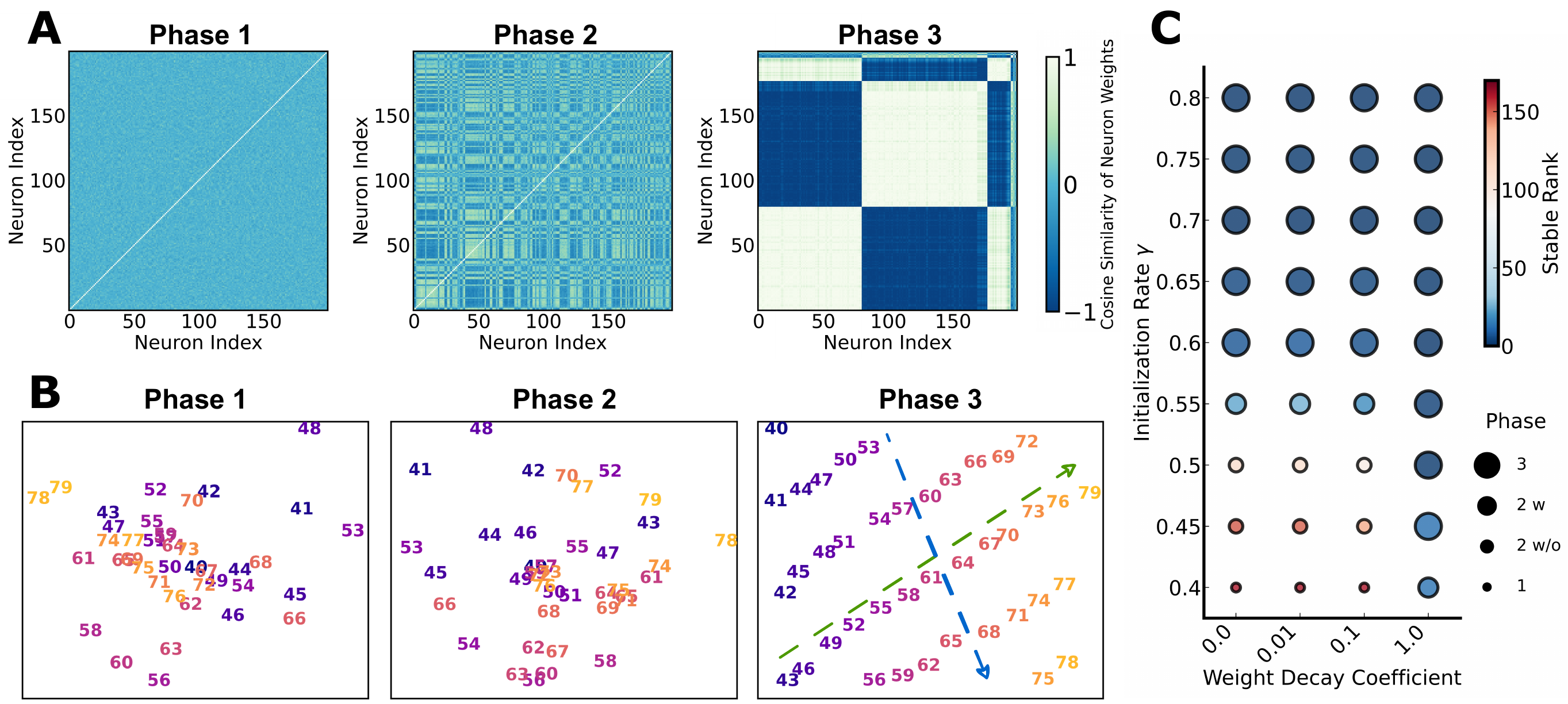}
\caption{\textbf{(A)} Cosine similarity between the input weights of neurons in the first layer’s query weight matrix (\(W^{Q(1)}\)) for each phase. The abscissa and ordinate both represent the neuron index. The matrices are computed under the settings where the weight decay coefficient is fixed at 0.01, and the initialization rate (\(\gamma\)) is set to 0.2, 0.5, and 0.8 for Phase 1, Phase 2, and Phase 3, respectively.  
\textbf{(B)} PCA visualization of the word embedding vectors for the same settings as in (A). Each number corresponds to a specific token, and its position represents the reduced-dimensional embedding obtained through PCA.  
\textbf{(C)} Bubble plot summarizing the stable rank of parameter matrices (\(W^{Q(1)}\)) across initialization rates (\(\gamma\)) and weight decay coefficients. The size of each bubble represents the phase index (w: with commutativity; w/o: without commutativity), with Phase 3 and Phase 2 distinguished based on whether the OOD accuracy exceeds 50\%, while other phase boundaries remain consistent with Fig.~\ref{fig:phases}. The color of each bubble represents the stable rank value, and all results are averaged over three independent trials.  }
\label{fig:rank}
\end{figure*}

\subsection{Condensation of Input Weights}

To investigate the phenomenon of parameter condensation, we analyze the cosine similarity between the input weights of neurons in the first layer’s query weight matrix (\( W^{Q(1)} \)). The similarity between the \( i \)-th and \( j \)-th neurons is calculated as 
\begin{equation*}
    \frac{W^{Q(1)}{[i,:]} \cdot W^{Q(1)}{[j,:]}}{||W^{Q(1)}{[i,:]}||_{2} ||W^{Q(1)}{[j,:]}||_{2}},
\end{equation*}
and the results are visualized in Fig.~\ref{fig:rank}A for each phase.  

In Phase 1, the input weights exhibit no significant condensation, with low cosine similarity between neurons. This indicates that the neurons are oriented independently in the weight space, corresponding to a high-complexity regime. In Phase 2, we observe mild condensation, where neurons begin clustering along a limited number of orientations, resulting in slight increases in similarity. Finally, in Phase 3, the input weights cluster strongly into a few isolated directions, forming distinct groups of neurons with high intra-group similarity. This pronounced condensation reflects a significant reduction in complexity, allowing the model to approximate a lower-dimensional structure while maintaining task performance.

\subsection{Structured Organization of the Word Embedding Matrix  }
To explore the structural properties of the model's word embedding matrix, we visualize the embedding space using PCA, as shown in Fig.~\ref{fig:rank}B. Each number represents the PCA-reduced position of the embedding vector corresponding to a specific token.  

In Phase 1 and Phase 2, the embedding vectors do not exhibit any clear structure, with token positions scattered irregularly in the reduced space. This lack of organization suggests that the model does not encode systematic relationships between tokens. However, in Phase 3, the PCA visualization reveals a highly structured pattern, with token embeddings arranged in an organized and regular manner. This structure reflects the model’s tendency to encode relationships in a systematic, rule-based fashion, enabling it to generalize efficiently while maintaining reduced complexity.  

It is worth noting that the relative ordinal relationship of the word embedding matrix in Phase 3 is not a simple numerical magnitude relationship of the corresponding tokens. This ordinal relationship may originate from the definition of the four single anchors, where the differences between the operations of any two single anchors can be obtained by the addition of the basic elements 3 and 4. This arrangement is consistent with the numbers being ordered with intervals of 3 (green arrow) and 4 (blue arrow) from two directions in the embedding space. 

\begin{figure*}
\centering
\includegraphics[width=0.9\linewidth]{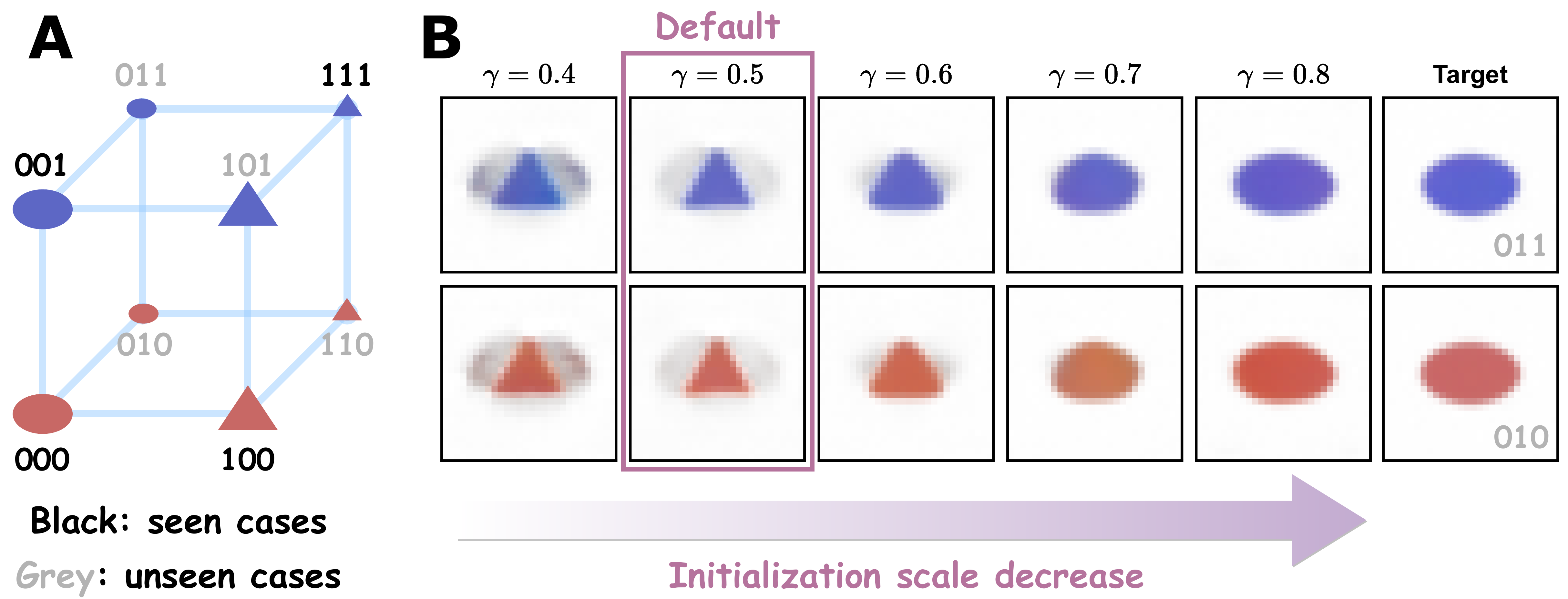}
\caption{\textbf{(A)} Structure of the Concept Graphs dataset. The dataset is organized into a cubic framework based on three core concept variables—color, shape, and size—each with specific values (\textit{e.g.}, red or blue for color, circle or triangle for shape, and large or small for size). Black nodes represent the sparse subset of training data used to identify the interesting failure mode. \textbf{(B)} Impact of varying initialization rate on compositional generalization for cases 011 and 010. As the initialization rate increases (corresponding to a decrease in the initialization scale), the model transitions from exhibiting the interesting failure mode to achieving robust compositional generalization. The purple box highlights the default initialization setting used in prior work.}
\label{fig:diffu}
\end{figure*}

\subsection{Unified Complexity Perspective via Stable Rank}
To unify the observed phenomena of condensation in input weights and structured organization in word embeddings, we analyze the complexity of parameter matrices across phases using \textbf{stable rank}~\cite{rudelson2007sampling,tropp2015introduction,vershynin2018high}, a widely adopted measure of matrix complexity. The stable rank is defined as:  
\[
R_{\text{stable}}(A) = \frac{\|A\|_F^2}{\|A\|_2^2},
\]  
where \( \|A\|_F \) denotes the Frobenius norm, representing the overall energy of the matrix, and \( \|A\|_2 \) denotes the spectral norm, capturing the largest singular value of the matrix. This ratio quantifies the ``effective dimensionality'' of the matrix, providing a normalized measure of complexity.  

In Fig.~\ref{fig:rank}C, we analyze the first-layer query weight matrices from the multi-head GPT-2 models introduced in Fig.~\ref{fig:phases}. The weights from all heads are merged into a single square matrix for each configuration. The horizontal and vertical axes represent the weight decay coefficient and initialization rate (\( \gamma \)), respectively. The size of each bubble indicates the phase index under the given configuration, where Phase 3 and Phase 2 are distinguished based on whether the OOD accuracy exceeds 50\%, with other phase boundaries consistent with Fig.~\ref{fig:phases}. The color of each bubble reflects the stable rank of the parameter matrix for that configuration. Each result is averaged by three random seeds.

As the model transitions through phases, the stable rank gradually decreases with increasing phase index, indicating a reduction in model complexity as the reasoning preference becomes more pronounced. Within the same phase, the stable rank values remain relatively consistent across different setups, further emphasizing the alignment between the stable rank and the model's phase-specific complexity characteristics. Additionally, in Appendix~\ref{detailed_condense}, we illustrate the degree of parameter condensation under various configurations, which is highly consistent with our stable rank findings.

The interplay between model complexity, internal mechanisms, and reasoning ability is crucial in shaping a Transformer's capacity for compositional generalization. Complexity control—achieved by adjusting initialization rates and weight decay coefficients—shapes the model’s parameter structure, resulting in neuron input weight condensation and a structured word embedding matrix. Neuron condensation streamlines input representations, facilitating the recognition of essential patterns, while structured embeddings align token representations with underlying compositional rules. These internal refinements enhance the model's reasoning capabilities by enabling it to focus on core compositional primitives rather than memorizing extensive input-output mappings.

\section{Further Verification on Realistic Tasks}\label{sec:real_task}

We validated the performance of models with different initialization scales and weight decay settings across a series of compositional and reasoning tasks. Below, we introduce each task and the corresponding results.

\textbf{Compositional diffusion tasks: Concept Graphs.} The Concept Graphs dataset~\cite{okawa2023compositional, parkemergence, yang2024dynamics, okawa2024compositional} provides a synthetic and structured framework designed to evaluate compositional generalization in conditional generative models. Built upon three core concept variables—color, shape, and size—each taking on specific values (\textit{e.g.}, red or blue for color, circle or triangle for shape, and large or small for size), the dataset organizes the relationships between different concept combinations in the form of a cube, as shown in Fig.~\ref{fig:diffu}A. Previous work identified a notable example of compositional generalization failure, which they referred to as an \textbf{\textit{interesting failure mode}}~\cite{okawa2023compositional}.  This failure occurs when the model is trained on a sparse subset of the data, specifically the black nodes in Fig.~\ref{fig:diffu}A (000, 100, 001, and 111). When tested on an unseen tuple like 010 (representing a small red circle), the model incorrectly generates a small triangle instead of the expected circle.  This demonstrates the model's inability to achieve compositional generalization under these conditions.

Building on our analyses of the impact of initialization on a model’s ability to learn compositional tasks, we train models for this task with varying initialization scales. We discover that when the initialization scale is smaller, the interesting failure mode disappears, and the model achieves robust compositional generalization. As illustrated in Fig.~\ref{fig:diffu}B, we gradually increase the initialization rate (decreasing the initialization scale) from left to right. The model’s outputs for the two unseen concept tuples (010 and 011) progressively transition from triangles to circles, aligning with the target shapes. The purple box highlights the initialization setting used in prior work~\cite{okawa2023compositional}, referred to as the default initialization. Notably, to replicate the exact behavior at \( \gamma = 0.5 \) (default setting) observed in prior work, we employ a uniform distribution for parameter initialization $\vtheta \sim \mathcal{U}(-1/d_{\text{in}}^{\gamma}, 1/d_{\text{in}}^{\gamma})$ in this experiment (PyTorch’s default \texttt{kaiming\_uniform} distribution, which is adopted in prior work). Consistent results are also observed when using a normal distribution for initialization.

\begin{figure*}[ht]
\centering
\includegraphics[width=0.98\linewidth]{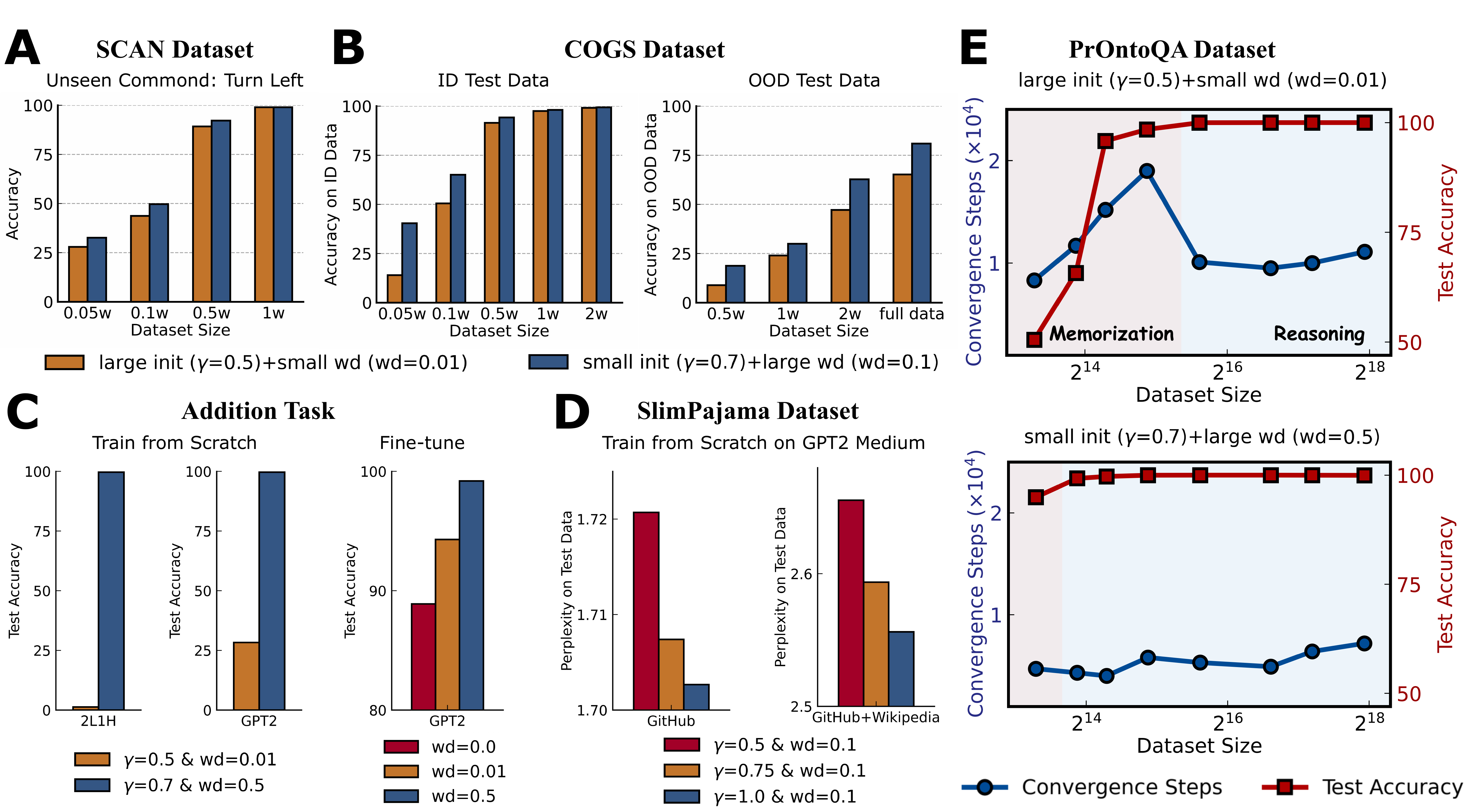}

\caption{\textbf{(A, B)} Performance comparison of models with different initialization scales and weight decay coefficients on compositional tasks. \textbf{(A)} For the SCAN task, we assess the generalization ability on composite commands that include the ``turn left" command. \textbf{(B)} For the COGS task, we evaluate (left panel) ID and (right panel) OOD generalization after training on the same dataset. Small initialization and large weight decay (blue) consistently outperform large initialization and small weight decay (orange) across different tasks and data scales. The parameters are initialized following a zero-mean normal distribution with a standard deviation of $d_{in}^{-\gamma}$. \textbf{(C)} Performance comparison of models with different initialization scales and weight decay coefficients on addition task. We use a case-based reasoning intervention experiment with a test set of \(a, b \in [400, 600]\) and the remaining data as the training set. For the ``Train from Scratch'' part, the parameters are initialized following a zero-mean normal distribution with a standard deviation of $d_{in}^{-\gamma}$. For the ``Fine-tune'' part, we use the pre-trained weights of the GPT-2 model provided by Hugging Face as the starting point for fine-tuning. \textbf{(D)} Performance comparison of models with different initialization scales and weight decay coefficients on the SlimPajama dataset. For the SlimPajama dataset, GPT-2 Medium models trained on GitHub and GitHub+Wikipedia data with small initialization scales consistently achieved lower perplexity. The parameters are initialized following a zero-mean normal distribution with a standard deviation of $d_{in}^{-\gamma}$. \textbf{(E)} Performance comparison of models with different initialization scales and weight decay coefficients on PrOntoQA. Top Panel: Convergence steps and test accuracy for large initialization~($\gamma=0.5$) and small weight decay~(WD = 0.01). Bottom Panel: Convergence steps and test accuracy for small initialization~($\gamma=0.7$) and large weight decay~(WD = 0.1). The parameters are initialized following a zero-mean normal distribution with a standard deviation of $d_{in}^{-\gamma}$.}
\label{fig:real_task}
\end{figure*}

\textbf{Compositional tasks: SCAN and COGS.} SCAN~\cite{lake2018generalization} and COGS~\cite{kim2020cogs} are classic compositional tasks with more natural language variance. For the SCAN dataset, we selected the ``Generalizing composition across primitive commands" task, where the ``turn left" command only appears in single-command mappings and is trained alongside other composite commands. We assess the model's generalization ability on composite commands that include the ``turn left" command. For the COGS dataset, we evaluate ID and OOD generalization after training on the same set. ID tests use data with different primitives in the same combinatorial patterns, while OOD tests use data following different combinatorial rules.

As shown in Figs.~\ref{fig:real_task}A, \ref{fig:real_task}B, we display the generalization performance of models with different initialization scales and weight decay coefficients across various data sizes. Small initialization and large weight decay consistently outperform large initialization and small weight decay across different task types and data scales. Notably, in the COGS task, even when the ID generalization of both settings (with 20k training data) reaches over 99\%, the difference in OOD generalization remains significant.

\textbf{Realistic tasks: Addition task and SlimPajama dataset.} Unlike traditional addition tasks, we use a case-based reasoning intervention experiment~\cite{hucase} to study the generalization of rule learning in the addition task. Specifically, we consider the setup: $a + b = c$, where $a, b \in [0, 999]$. We use $a, b \in [400, 600]$ as the test set and the remaining data as the training set. This construction prevents the model from simply mimicking training data similar to the test set. We trained a simple 2-layer 1-head model and a GPT-2 model. As shown in Fig.~\ref{fig:real_task}C, regardless of model size and learning mode, small initialization scales (or large weight decay coefficients) generally lead to good rule generalization, while large initialization scales (or small weight decay coefficients) fail to generalize perfectly.

For the SlimPajama dataset~\cite{cerebras2023slimpajama}, we used two data compositions: the GitHub section and the GitHub+Wikipedia section. We trained GPT-2 Medium models with different initializations on both datasets for 40B tokens. As shown in Fig.~\ref{fig:real_task}D, for both data compositions, smaller initialization scales consistently achieved lower perplexity. 

\textbf{Reasoning tasks: PrOntoQA.} 
PrOntoQA~\cite{saparovlanguage} is a synthetic multi-step reasoning dataset where each data point assigns hierarchical relationships among objects and requires the model to determine whether a multi-step reasoning chain is correct. During testing, we only evaluate the model's accuracy in judging hierarchical relationships. Thus, the model's random guessing accuracy is 50\%. 

\begin{figure*}[ht]
\centering
\includegraphics[width=0.98\linewidth]{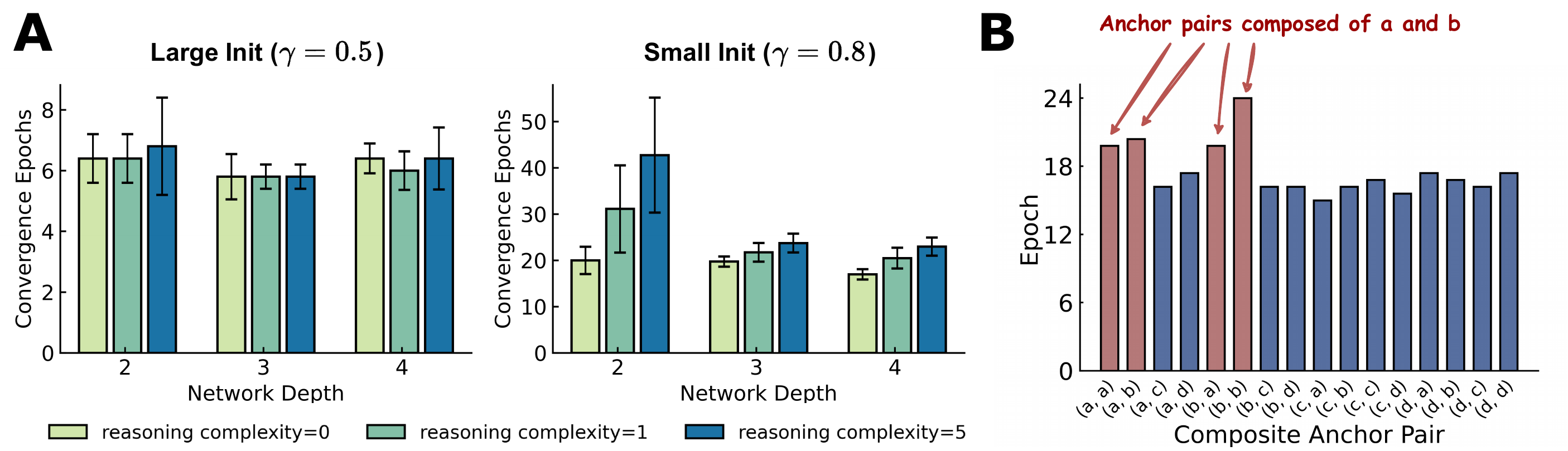}
\caption{\textbf{(A)} Training speed with different initialization scales when facing different data complexity. Abscissa: the model depth. Ordinate: the number of epochs required for the model's training accuracy to reach 100\%. Color: the inferential complexity. We conduct 9 random trials for each setting of different initializations, depths, and data inferential complexities, and take the average of the required number of epochs. Left Panel: Large initialization (initialization rate $\gamma=0.5$). Right Panel: Small initialization (initialization rate $\gamma=0.8$). \textbf{(B)} The number of epochs required for different anchor pairs to reach 60\% accuracy on their corresponding training data. The anchor pairs ($a$, $b$), ($b$, $a$), ($a$, $a$), and ($b$, $b$) are highlighted in red because they are significantly affected by the non-inferential solutions ($a$, $b$) and ($b$, $a$).}
\label{fig:data_complex}
\end{figure*}

Fig.~\ref{fig:real_task}E illustrates the convergence rates and generalization errors with respect to data scale for models with large initialization (and small weight decay coefficients, top panel) and small initialization (and large weight decay coefficients, bottom panel). An interesting phenomenon is observed for models with large initialization (small weight decay coefficients): as the data size increases, the convergence rate first decreases and then increases. When the data size is small, the model tends to fit the data through memorization. Therefore, as the data size increases, the training difficulty increases (\textit{i.e.}, the training speed slows down), and the model's generalization ability is poor. As the data size grows further, the model, constrained by its complexity, can no longer memorize all the data and thus shifts to fitting the data through reasoning. This leads to an increase in fitting speed and results in better generalization. In contrast, models with small initialization (large weight decay coefficients) inherently prefer to fit the data through reasoning, leading to faster convergence and better generalization at the same data scale compared to models with large initialization (small weight decay coefficients).

\section{Prediction of Mechanisms underlying Different Solutions}\label{sec:pred}
As the data complexity increases, neural networks usually require more training steps to fit the data~\cite{arpit2017closer,rahaman2018spectral,xu2019frequency,xu_training_2019}. Based on our analysis of the mechanisms underlying different solutions, we can predict that models with different initialization scales will exhibit varying degrees of difficulty when learning data of differing complexities. To create datasets with varying levels of complexity, we deliberately modify certain composite mappings while ensuring that anchor pairs with symmetrical relationships remain consistent (\textit{i.e.}, $f(x;a_i, a_j) = f(x;a_j, a_i) \neq g(g(x;a_i);a_j)$). We define the \textbf{reasoning complexity} as the number of anchor pair groups\footnote{Two anchor pairs with symmetrical relationships are considered to be in the same group.} that do not satisfy the compositional rule. This reasoning complexity captures the variety of data in the dataset that violate reasoning rules and necessitate memory-based mappings.  A higher reasoning complexity indicates more intricate relationships and patterns that simple reasoning rules cannot easily capture. Consequently, the model must rely more on memorizing specific data mappings rather than applying general reasoning principles.

It is anticipated that the model of small initialization, \textit{i.e.}, fitting data with as low complexity as possible, should use more training steps to fit data with larger reasoning complexity, while the model with large initialization, \textit{i.e.}, easily memorizing the mappings of the symmetric solution, should use similar training steps to fit data with different reasoning complexity.

To measure the training steps required to fit data, we use the number of epochs required for the model's training loss to reach $5 \times 10^{-2}$. We conduct 9 random trials for each setting.
As shown in Fig.~\ref{fig:data_complex}A, models that tend to learn data with memorizing solutions use roughly the same training steps for data with different reasoning complexity (the left panel of  Fig.~\ref{fig:data_complex}A), while models that tend to learn data with reasoning solutions indeed use more training steps for data with larger reasoning complexity (the right panel of  Fig.~\ref{fig:data_complex}A). This prediction convincingly backs up our analysis of why different initialization scales can lead to different solutions.

We further demonstrate the fitting speed of different anchor pairs when the reasoning complexity is one. In this experiment, we simultaneously set the anchor pairs ($a$, $b$) and ($b$, $a$) as the same mappings violating the compositional rules. As shown in Fig.~\ref{fig:data_complex}B, we present the number of epochs required for 16 types of anchor pairs to achieve 60\% accuracy on their corresponding training data. We highlight the four types of anchor pairs ($a$, $b$), ($b$, $a$), ($a$, $a$), and ($b$, $b$) in red (since ($a$, $b$) and ($b$, $a$) violate the compositional rules, the above four types of anchor pairs will be significantly affected). It can be observed that the highlighted anchor pair types exhibit notably slower training speeds, further verifying that the presence of mappings that violate the compositional rules reduces the training speed of models with small initialization.

From another perspective, this also allows us to discover compositional rule-violating mappings (mostly noisy data) in the dataset. For a set of data with unknown distribution, we can differentiate to a certain extent by studying the convergence rate of different types of data using models with small initialization.

\section{Discussion} \label{sec:dis}

\textbf{Conclusion. }In this work, we examined the internal mechanisms of Transformer models when tackling compositional tasks, particularly how complexity control—through parameter initialization scales and weight decay—shapes the model’s reasoning capabilities and generalization performance. By introducing anchor functions as controlled, interpretable benchmarks, we were able to precisely delineate the conditions under which Transformers transition from memory-based solutions, which merely memorize input-output mappings, to reasoning-based solutions that effectively capture primitive-level rules.

Our analyses revealed that small initialization scales and larger weight decay coefficients lead models toward solutions that are inherently more structured, systematic, and capable of handling unseen compositional combinations. Such solutions were shown to exhibit a lower complexity bias, allowing the model to both identify and exploit core compositional primitives. Through a series of mechanism analyses, including masking strategies and complexity quantification via measures like stable rank, we demonstrated how differences in complexity phases correspond to distinct internal working mechanisms.

We validated the generality of these findings across multiple real-world tasks, including diffusion-based image generation and various natural language benchmarks. Models that embraced a reasoning-based approach consistently achieved improved OOD generalization, underscoring the practical value of complexity control. Furthermore, the observed relationships between complexity parameters and the time needed for training on data of varying reasoning complexity highlight the predictive power of our framework. These insights enable a more principled approach to anticipating model behavior, diagnosing performance bottlenecks, and guiding architectural or training choices for improved compositional reasoning.

\textbf{Limitation and Future Work. }One key limitation of our study is that, although complexity control techniques have been validated across multiple real-world tasks, our analysis of internal mechanisms was confined to synthetic data and smaller-scale models. This constraint may not fully capture the intricacies and challenges inherent in large-scale, real-world datasets and models. Furthermore, enhancing the model's preference for reasoning may partially compromise its memorization capabilities. This trade-off poses a challenge, as reduced memorization can limit the model's ability to recall specific information that may be critical for certain tasks. To address this issue, we propose leveraging Retrieval-Augmented Generation (RAG) techniques to compensate for the diminished memorization. By integrating RAG, we can offload memory-intensive tasks to external retrieval systems, thereby allowing the complexity control mechanisms to focus exclusively on enhancing the model's reasoning abilities without compromising its overall performance.

In future work, we aim to extend our investigation to larger models and diverse real-world datasets to bridge the gap between our theoretical understanding and practical applications. This could involve leveraging Mixture of Experts (MoEs) to design networks with varying initialization scales for different expert models, employing RAG techniques to enhance the network's memory capabilities, and exploring improved training strategies that utilize complexity control methods to ensure the main network focuses on reasoning abilities while auxiliary components manage memorization tasks.

\section*{Acknowledgments}
This work is sponsored by the National Key R\&D Program of China Grant No. 2022YFA1008200, the National Natural Science Foundation of China Grant No. 92270001(Z. X.), 12371511 (Z. X.), 12422119 (Z. X.), 12101402 (Y. Z.), the Lingang Laboratory Grant No. LG-QS-202202-08 (Y. Z.), Shanghai Municipal of Science and Technology Major Project No. 2021SHZDZX0102 (Z. X., Y. Z.), and the HPC of School of Mathematical Sciences and the Student Innovation Center, and the Siyuan-1 cluster supported by the Center for High Performance Computing at Shanghai Jiao Tong University, Key Laboratory of Marine Intelligent Equipment and System, Ministry of Education, P.R. China. This work was partially supported by SJTU Kunpeng\&Ascend Center of Excellence. 

\bibliographystyle{IEEEtran}
\bibliography{neurips_2024}

\begin{thebibliography}{10}
\providecommand{\url}[1]{#1}
\csname url@samestyle\endcsname
\providecommand{\newblock}{\relax}
\providecommand{\bibinfo}[2]{#2}
\providecommand{\BIBentrySTDinterwordspacing}{\spaceskip=0pt\relax}
\providecommand{\BIBentryALTinterwordstretchfactor}{4}
\providecommand{\BIBentryALTinterwordspacing}{\spaceskip=\fontdimen2\font plus
\BIBentryALTinterwordstretchfactor\fontdimen3\font minus \fontdimen4\font\relax}
\providecommand{\BIBforeignlanguage}[2]{{%
\expandafter\ifx\csname l@#1\endcsname\relax
\typeout{** WARNING: IEEEtran.bst: No hyphenation pattern has been}%
\typeout{** loaded for the language `#1'. Using the pattern for}%
\typeout{** the default language instead.}%
\else
\language=\csname l@#1\endcsname
\fi
#2}}
\providecommand{\BIBdecl}{\relax}
\BIBdecl

\bibitem{vaswani2017attention}
A.~Vaswani, N.~Shazeer, N.~Parmar, J.~Uszkoreit, L.~Jones, A.~N. Gomez, {\L}.~Kaiser, and I.~Polosukhin, ``Attention is all you need,'' in \emph{Advances in neural information processing systems}, 2017, pp. 5998--6008.

\bibitem{achiam2023gpt}
J.~Achiam, S.~Adler, S.~Agarwal, L.~Ahmad, I.~Akkaya, F.~L. Aleman, D.~Almeida, J.~Altenschmidt, S.~Altman, S.~Anadkat \emph{et~al.}, ``Gpt-4 technical report,'' \emph{arXiv preprint arXiv:2303.08774}, 2023.

\bibitem{brown2020language}
T.~Brown, B.~Mann, N.~Ryder, M.~Subbiah, J.~D. Kaplan, P.~Dhariwal, A.~Neelakantan, P.~Shyam, G.~Sastry, A.~Askell \emph{et~al.}, ``Language models are few-shot learners,'' \emph{Advances in neural information processing systems}, vol.~33, pp. 1877--1901, 2020.

\bibitem{choi2021chatgpt}
J.~H. Choi, K.~E. Hickman, A.~B. Monahan, and D.~Schwarcz, ``Chatgpt goes to law school,'' \emph{J. Legal Educ.}, vol.~71, p. 387, 2021.

\bibitem{teubner2023welcome}
T.~Teubner, C.~M. Flath, C.~Weinhardt, W.~van~der Aalst, and O.~Hinz, ``Welcome to the era of chatgpt et al. the prospects of large language models,'' \emph{Business \& Information Systems Engineering}, vol.~65, no.~2, pp. 95--101, 2023.

\bibitem{bubeck2023sparks}
S.~Bubeck, V.~Chandrasekaran, R.~Eldan, J.~Gehrke, E.~Horvitz, E.~Kamar, P.~Lee, Y.~T. Lee, Y.~Li, S.~Lundberg \emph{et~al.}, ``Sparks of artificial general intelligence: Early experiments with gpt-4,'' \emph{arXiv preprint arXiv:2303.12712}, 2023.

\bibitem{marcus2003algebraic}
G.~F. Marcus, \emph{The algebraic mind: Integrating connectionism and cognitive science}.\hskip 1em plus 0.5em minus 0.4em\relax MIT press, 2003.

\bibitem{smolensky2022neurocompositional}
P.~Smolensky, R.~McCoy, R.~Fernandez, M.~Goldrick, and J.~Gao, ``Neurocompositional computing: From the central paradox of cognition to a new generation of ai systems,'' \emph{AI Magazine}, vol.~43, no.~3, pp. 308--322, 2022.

\bibitem{fodor1988connectionism}
J.~A. Fodor and Z.~W. Pylyshyn, ``Connectionism and cognitive architecture: A critical analysis,'' \emph{Cognition}, vol.~28, no. 1-2, pp. 3--71, 1988.

\bibitem{keysers2019measuring}
D.~Keysers, N.~Sch{\"a}rli, N.~Scales, H.~Buisman, D.~Furrer, S.~Kashubin, N.~Momchev, D.~Sinopalnikov, L.~Stafiniak, T.~Tihon \emph{et~al.}, ``Measuring compositional generalization: A comprehensive method on realistic data,'' in \emph{International Conference on Learning Representations}, 2019.

\bibitem{yu2020assessing}
L.~Yu and A.~Ettinger, ``Assessing phrasal representation and composition in transformers,'' in \emph{Proceedings of the 2020 Conference on Empirical Methods in Natural Language Processing (EMNLP)}, 2020, pp. 4896--4907.

\bibitem{hupkes2020compositionality}
D.~Hupkes, V.~Dankers, M.~Mul, and E.~Bruni, ``Compositionality decomposed: How do neural networks generalise?'' \emph{Journal of Artificial Intelligence Research}, vol.~67, pp. 757--795, 2020.

\bibitem{press2023measuring}
O.~Press, M.~Zhang, S.~Min, L.~Schmidt, N.~A. Smith, and M.~Lewis, ``Measuring and narrowing the compositionality gap in language models,'' in \emph{Findings of the Association for Computational Linguistics: EMNLP 2023}, 2023, pp. 5687--5711.

\bibitem{kim2020cogs}
N.~Kim and T.~Linzen, ``Cogs: A compositional generalization challenge based on semantic interpretation,'' in \emph{Proceedings of the 2020 conference on empirical methods in natural language processing (emnlp)}, 2020, pp. 9087--9105.

\bibitem{wang2024grokked}
B.~Wang, X.~Yue, Y.~Su, and H.~Sun, ``Grokked transformers are implicit reasoners: A mechanistic journey to the edge of generalization,'' \emph{arXiv preprint arXiv:2405.15071}, 2024.

\bibitem{zhang2024initialization}
Z.~Zhang, P.~Lin, Z.~Wang, Y.~Zhang, and Z.-Q.~J. Xu, ``Initialization is critical to whether transformers fit composite functions by reasoning or memorizing,'' in \emph{The Thirty-eighth Annual Conference on Neural Information Processing Systems}, 2024.

\bibitem{fu2022does}
Y.~Fu, H.~Peng, and T.~Khot, ``How does gpt obtain its ability? tracing emergent abilities of language models to their sources,'' \emph{Yao Fu’s Notion}, 2022.

\bibitem{wei2022emergent}
J.~Wei, Y.~Tay, R.~Bommasani, C.~Raffel, B.~Zoph, S.~Borgeaud, D.~Yogatama, M.~Bosma, D.~Zhou, D.~Metzler \emph{et~al.}, ``Emergent abilities of large language models,'' \emph{arXiv preprint arXiv:2206.07682}, 2022.

\bibitem{srivastava2022beyond}
A.~Srivastava, A.~Rastogi, A.~Rao, A.~A.~M. Shoeb, A.~Abid, A.~Fisch, A.~R. Brown, A.~Santoro, A.~Gupta, A.~Garriga-Alonso \emph{et~al.}, ``Beyond the imitation game: Quantifying and extrapolating the capabilities of language models,'' \emph{arXiv preprint arXiv:2206.04615}, 2022.

\bibitem{csordas2021neural}
R.~Csord{\'a}s, K.~Irie, and J.~Schmidhuber, ``The neural data router: Adaptive control flow in transformers improves systematic generalization,'' \emph{arXiv preprint arXiv:2110.07732}, 2021.

\bibitem{dziri2024faith}
N.~Dziri, X.~Lu, M.~Sclar, X.~L. Li, L.~Jiang, B.~Y. Lin, S.~Welleck, P.~West, C.~Bhagavatula, R.~Le~Bras \emph{et~al.}, ``Faith and fate: Limits of transformers on compositionality,'' \emph{Advances in Neural Information Processing Systems}, vol.~36, 2024.

\bibitem{hupkes2018learning}
D.~Hupkes, A.~Singh, K.~Korrel, G.~Kruszewski, and E.~Bruni, ``Learning compositionally through attentive guidance,'' \emph{arXiv preprint arXiv:1805.09657}, 2018.

\bibitem{lepori2023break}
M.~A. Lepori, T.~Serre, and E.~Pavlick, ``Break it down: Evidence for structural compositionality in neural networks,'' \emph{arXiv preprint arXiv:2301.10884}, 2023.

\bibitem{okawa2023compositional}
M.~Okawa, E.~S. Lubana, R.~P. Dick, and H.~Tanaka, ``Compositional abilities emerge multiplicatively: Exploring diffusion models on a synthetic task,'' \emph{arXiv preprint arXiv:2310.09336}, 2023.

\bibitem{yun2022vision}
T.~Yun, U.~Bhalla, E.~Pavlick, and C.~Sun, ``Do vision-language pretrained models learn composable primitive concepts?'' \emph{arXiv preprint arXiv:2203.17271}, 2022.

\bibitem{wang2024towards}
Z.~Wang, Y.~Wang, Z.~Zhang, Z.~Zhou, H.~Jin, T.~Hu, J.~Sun, Z.~Li, Y.~Zhang, and Z.-Q.~J. Xu, ``Towards understanding how transformer perform multi-step reasoning with matching operation,'' \emph{arXiv preprint arXiv:2405.15302}, 2024.

\bibitem{csordas2022ctl}
R.~Csord{\'a}s, K.~Irie, and J.~Schmidhuber, ``Ctl++: Evaluating generalization on never-seen compositional patterns of known functions, and compatibility of neural representations,'' \emph{arXiv preprint arXiv:2210.06350}, 2022.

\bibitem{ramesh2023capable}
R.~Ramesh, M.~Khona, R.~P. Dick, H.~Tanaka, and E.~S. Lubana, ``How capable can a transformer become? a study on synthetic, interpretable tasks,'' \emph{arXiv preprint arXiv:2311.12997}, 2023.

\bibitem{liu2022transformers}
B.~Liu, J.~T. Ash, S.~Goel, A.~Krishnamurthy, and C.~Zhang, ``Transformers learn shortcuts to automata,'' \emph{arXiv preprint arXiv:2210.10749}, 2022.

\bibitem{wei2022chain}
J.~Wei, X.~Wang, D.~Schuurmans, M.~Bosma, E.~Chi, Q.~Le, and D.~Zhou, ``Chain of thought prompting elicits reasoning in large language models,'' \emph{arXiv preprint arXiv:2201.11903}, 2022.

\bibitem{creswell2022selection}
A.~Creswell, M.~Shanahan, and I.~Higgins, ``Selection-inference: Exploiting large language models for interpretable logical reasoning,'' \emph{arXiv preprint arXiv:2205.09712}, 2022.

\bibitem{creswell2022faithful}
A.~Creswell and M.~Shanahan, ``Faithful reasoning using large language models,'' \emph{arXiv preprint arXiv:2208.14271}, 2022.

\bibitem{wang2024improving}
M.~Wang, H.~He, J.~Wang, Z.~Wang, G.~Huang, F.~Xiong, Z.~Li, L.~Wu \emph{et~al.}, ``Improving generalization and convergence by enhancing implicit regularization,'' \emph{arXiv preprint arXiv:2405.20763}, 2024.

\bibitem{wang2024understanding}
M.~Wang \emph{et~al.}, ``Understanding the expressive power and mechanisms of transformer for sequence modeling,'' \emph{arXiv preprint arXiv:2402.00522}, 2024.

\bibitem{wang2023label}
L.~Wang, L.~Li, D.~Dai, D.~Chen, H.~Zhou, F.~Meng, J.~Zhou, and X.~Sun, ``Label words are anchors: An information flow perspective for understanding in-context learning,'' \emph{arXiv preprint arXiv:2305.14160}, 2023.

\bibitem{cao2024graphinsight}
Y.~Cao, S.~Han, Z.~Gao, Z.~Ding, X.~Xie, and S.~K. Zhou, ``Graphinsight: Unlocking insights in large language models for graph structure understanding,'' \emph{arXiv preprint arXiv:2409.03258}, 2024.

\bibitem{zhang2024anchor}
Z.~Zhang, Z.~Wang, J.~Yao, Z.~Zhou, X.~Li, W.~{E}, and Z.-Q.~J. Xu, ``Anchor function: a type of benchmark functions for studying language models,'' \emph{arXiv preprint arXiv:2401.08309}, 2024.

\bibitem{marcus2022very}
G.~Marcus, E.~Davis, and S.~Aaronson, ``A very preliminary analysis of dall-e 2,'' \emph{arXiv preprint arXiv:2204.13807}, 2022.

\bibitem{leivada2023dall}
E.~Leivada, E.~Murphy, and G.~Marcus, ``Dall{\textperiodcentered} e 2 fails to reliably capture common syntactic processes,'' \emph{Social Sciences \& Humanities Open}, vol.~8, no.~1, p. 100648, 2023.

\bibitem{conwell2022testing}
C.~Conwell and T.~Ullman, ``Testing relational understanding in text-guided image generation,'' \emph{arXiv preprint arXiv:2208.00005}, 2022.

\bibitem{gokhale2022benchmarking}
T.~Gokhale, H.~Palangi, B.~Nushi, V.~Vineet, E.~Horvitz, E.~Kamar, C.~Baral, and Y.~Yang, ``Benchmarking spatial relationships in text-to-image generation,'' \emph{arXiv preprint arXiv:2212.10015}, 2022.

\bibitem{du2023reduce}
Y.~Du, C.~Durkan, R.~Strudel, J.~B. Tenenbaum, S.~Dieleman, R.~Fergus, J.~Sohl-Dickstein, A.~Doucet, and W.~S. Grathwohl, ``Reduce, reuse, recycle: Compositional generation with energy-based diffusion models and mcmc,'' in \emph{International conference on machine learning}.\hskip 1em plus 0.5em minus 0.4em\relax PMLR, 2023, pp. 8489--8510.

\bibitem{liu2022compositional}
N.~Liu, S.~Li, Y.~Du, A.~Torralba, and J.~B. Tenenbaum, ``Compositional visual generation with composable diffusion models,'' in \emph{European Conference on Computer Vision}.\hskip 1em plus 0.5em minus 0.4em\relax Springer, 2022, pp. 423--439.

\bibitem{fengtraining}
W.~Feng, X.~He, T.-J. Fu, V.~Jampani, A.~R. Akula, P.~Narayana, S.~Basu, X.~E. Wang, and W.~Y. Wang, ``Training-free structured diffusion guidance for compositional text-to-image synthesis,'' in \emph{The Eleventh International Conference on Learning Representations}.

\bibitem{okawa2024compositional}
M.~Okawa, E.~S. Lubana, R.~Dick, and H.~Tanaka, ``Compositional abilities emerge multiplicatively: Exploring diffusion models on a synthetic task,'' \emph{Advances in Neural Information Processing Systems}, vol.~36, 2024.

\bibitem{liang2024diffusion}
Q.~Liang, Z.~Liu, M.~Ostrow, and I.~Fiete, ``How diffusion models learn to factorize and compose,'' \emph{arXiv preprint arXiv:2408.13256}, 2024.

\bibitem{yang2024dynamics}
Y.~Yang, C.~F. Park, E.~S. Lubana, M.~Okawa, W.~Hu, and H.~Tanaka, ``Dynamics of concept learning and compositional generalization,'' \emph{arXiv preprint arXiv:2410.08309}, 2024.

\bibitem{parkemergence}
C.~F. Park, M.~Okawa, A.~Lee, E.~S. Lubana, and H.~Tanaka, ``Emergence of hidden capabilities: Exploring learning dynamics in concept space,'' in \emph{The Thirty-eighth Annual Conference on Neural Information Processing Systems}.

\bibitem{arora2019exact}
S.~Arora, S.~S. Du, W.~Hu, Z.~Li, R.~R. Salakhutdinov, and R.~Wang, ``On exact computation with an infinitely wide neural net,'' in \emph{Advances in Neural Information Processing Systems}, 2019, pp. 8141--8150.

\bibitem{chizat_global_2018}
L.~Chizat and F.~Bach, ``On the {Global} {Convergence} of {Gradient} {Descent} for {Over}-parameterized {Models} using {Optimal} {Transport},'' in \emph{Advances in {Neural} {Information} {Processing} {Systems} 31}, 2018, pp. 3036--3046.

\bibitem{zhang_type_2019}
Y.~Zhang, Z.-Q.~J. Xu, T.~Luo, and Z.~Ma, ``A type of generalization error induced by initialization in deep neural networks,'' \emph{arXiv:1905.07777 [cs, stat]}, 2019.

\bibitem{e2020comparative}
W.~E, C.~Ma, and L.~Wu, ``A comparative analysis of optimization and generalization properties of two-layer neural network and random feature models under gradient descent dynamics.'' \emph{Sci. China Math.}, vol.~63, 2020.

\bibitem{jacot_neural_2018}
A.~Jacot, F.~Gabriel, and C.~Hongler, ``Neural {Tangent} {Kernel}: {Convergence} and {Generalization} in {Neural} {Networks},'' in \emph{Advances in {Neural} {Information} {Processing} {Systems} 31}, 2018, pp. 8571--8580.

\bibitem{mei_mean_2018}
S.~Mei, A.~Montanari, and P.-M. Nguyen, ``A mean field view of the landscape of two-layer neural networks,'' \emph{Proceedings of the National Academy of Sciences}, vol. 115, no.~33, pp. E7665--E7671, 2018.

\bibitem{rotskoff_parameters_2018}
G.~Rotskoff and E.~Vanden-Eijnden, ``Parameters as interacting particles: long time convergence and asymptotic error scaling of neural networks,'' in \emph{Advances in {Neural} {Information} {Processing} {Systems} 31}, 2018, pp. 7146--7155.

\bibitem{sirignano_mean_2020}
J.~Sirignano and K.~Spiliopoulos, ``Mean field analysis of neural networks: {A} central limit theorem,'' \emph{Stochastic Processes and their Applications}, vol. 130, no.~3, pp. 1820--1852, 2020.

\bibitem{williams_gradient_2019}
\BIBentryALTinterwordspacing
F.~Williams, M.~Trager, C.~T. Silva, D.~Panozzo, D.~Zorin, and J.~Bruna, ``Gradient dynamics of shallow univariate relu networks,'' \emph{CoRR}, vol. abs/1906.07842, 2019. [Online]. Available: \url{http://arxiv.org/abs/1906.07842}
\BIBentrySTDinterwordspacing

\bibitem{luo2021phase}
T.~Luo, Z.-Q.~J. Xu, Z.~Ma, and Y.~Zhang, ``Phase diagram for two-layer relu neural networks at infinite-width limit,'' \emph{Journal of Machine Learning Research}, vol.~22, no.~71, pp. 1--47, 2021.

\bibitem{zhou2022empirical}
H.~Zhou, Q.~Zhou, Z.~Jin, T.~Luo, Y.~Zhang, and Z.-Q.~J. Xu, ``Empirical phase diagram for three-layer neural networks with infinite width,'' \emph{Advances in Neural Information Processing Systems}, 2022.

\bibitem{zhang2022linear}
Y.~Zhang, Z.~Zhang, L.~Zhang, Z.~Bai, T.~Luo, and Z.-Q.~J. Xu, ``Linear stability hypothesis and rank stratification for nonlinear models,'' \emph{arXiv preprint arXiv:2211.11623}, 2022.

\bibitem{zhang2023loss}
Z.~Zhang and Z.-Q.~J. Xu, ``Loss spike in training neural networks,'' \emph{arXiv preprint arXiv:2305.12133}, 2023.

\bibitem{zhang2023stochastic}
Z.~Zhang, Y.~Li, T.~Luo, and Z.-Q.~J. Xu, ``Stochastic modified equations and dynamics of dropout algorithm,'' \emph{arXiv preprint arXiv:2305.15850}, 2023.

\bibitem{zhang2024implicit}
Z.~Zhang and Z.-Q.~J. Xu, ``Implicit regularization of dropout,'' \emph{IEEE Transactions on Pattern Analysis and Machine Intelligence}, 2024.

\bibitem{power2022grokking}
A.~Power, Y.~Burda, H.~Edwards, I.~Babuschkin, and V.~Misra, ``Grokking: Generalization beyond overfitting on small algorithmic datasets,'' \emph{arXiv preprint arXiv:2201.02177}, 2022.

\bibitem{gopalani2024transformers}
P.~Gopalani, E.~S. Lubana, and W.~Hu, ``How do transformers fill in the blanks? a case study on matrix completion,'' in \emph{ICML 2024 Workshop on Mechanistic Interpretability}, 2024.

\bibitem{liu2022omnigrok}
Z.~Liu, E.~J. Michaud, and M.~Tegmark, ``Omnigrok: Grokking beyond algorithmic data,'' in \emph{The Eleventh International Conference on Learning Representations}, 2022.

\bibitem{huang2020improving}
X.~S. Huang, F.~Perez, J.~Ba, and M.~Volkovs, ``Improving transformer optimization through better initialization,'' in \emph{International Conference on Machine Learning}.\hskip 1em plus 0.5em minus 0.4em\relax PMLR, 2020, pp. 4475--4483.

\bibitem{liu2020understanding}
L.~Liu, X.~Liu, J.~Gao, W.~Chen, and J.~Han, ``Understanding the difficulty of training transformers,'' \emph{arXiv preprint arXiv:2004.08249}, 2020.

\bibitem{trockman2023mimetic}
A.~Trockman and J.~Z. Kolter, ``Mimetic initialization of self-attention layers,'' in \emph{International Conference on Machine Learning}.\hskip 1em plus 0.5em minus 0.4em\relax PMLR, 2023, pp. 34\,456--34\,468.

\bibitem{wang2024deepnet}
H.~Wang, S.~Ma, L.~Dong, S.~Huang, D.~Zhang, and F.~Wei, ``Deepnet: Scaling transformers to 1,000 layers,'' \emph{IEEE Transactions on Pattern Analysis and Machine Intelligence}, 2024.

\bibitem{zhang2019improving}
B.~Zhang, I.~Titov, and R.~Sennrich, ``Improving deep transformer with depth-scaled initialization and merged attention,'' \emph{arXiv preprint arXiv:1908.11365}, 2019.

\bibitem{zhu2021gradinit}
C.~Zhu, R.~Ni, Z.~Xu, K.~Kong, W.~R. Huang, and T.~Goldstein, ``Gradinit: Learning to initialize neural networks for stable and efficient training,'' \emph{Advances in Neural Information Processing Systems}, vol.~34, pp. 16\,410--16\,422, 2021.

\bibitem{radford2019language}
A.~Radford, J.~Wu, R.~Child, D.~Luan, D.~Amodei, I.~Sutskever \emph{et~al.}, ``Language models are unsupervised multitask learners,'' \emph{OpenAI blog}, vol.~1, no.~8, p.~9, 2019.

\bibitem{zhou2021towards}
H.~Zhou, Q.~Zhou, T.~Luo, Y.~Zhang, and Z.-Q.~J. Xu, ``Towards understanding the condensation of neural networks at initial training,'' \emph{arXiv preprint arXiv:2105.11686}, 2021.

\bibitem{zhang2021embedding}
Y.~Zhang, Z.~Zhang, T.~Luo, and Z.~J. Xu, ``Embedding principle of loss landscape of deep neural networks,'' \emph{Advances in Neural Information Processing Systems}, vol.~34, pp. 14\,848--14\,859, 2021.

\bibitem{zhang2022embedding}
Y.~Zhang, Y.~Li, Z.~Zhang, T.~Luo, and Z.-Q.~J. Xu, ``Embedding principle: a hierarchical structure of loss landscape of deep neural networks,'' \emph{Journal of Machine Learning vol}, vol.~1, pp. 1--45, 2022.

\bibitem{rudelson2007sampling}
M.~Rudelson and R.~Vershynin, ``Sampling from large matrices: An approach through geometric functional analysis,'' \emph{Journal of the ACM (JACM)}, vol.~54, no.~4, pp. 21--es, 2007.

\bibitem{tropp2015introduction}
J.~A. Tropp \emph{et~al.}, ``An introduction to matrix concentration inequalities,'' \emph{Foundations and Trends{\textregistered} in Machine Learning}, vol.~8, no. 1-2, pp. 1--230, 2015.

\bibitem{vershynin2018high}
R.~Vershynin, \emph{High-dimensional probability: An introduction with applications in data science}.\hskip 1em plus 0.5em minus 0.4em\relax Cambridge university press, 2018, vol.~47.

\bibitem{lake2018generalization}
B.~Lake and M.~Baroni, ``Generalization without systematicity: On the compositional skills of sequence-to-sequence recurrent networks,'' in \emph{International conference on machine learning}.\hskip 1em plus 0.5em minus 0.4em\relax PMLR, 2018, pp. 2873--2882.

\bibitem{hucase}
Y.~Hu, X.~Tang, H.~Yang, and M.~Zhang, ``Case-based or rule-based: How do transformers do the math?'' in \emph{Forty-first International Conference on Machine Learning}, 2024.

\bibitem{cerebras2023slimpajama}
\BIBentryALTinterwordspacing
D.~Soboleva, F.~Al-Khateeb, R.~Myers, J.~R. Steeves, J.~Hestness, and N.~Dey, ``{SlimPajama: A 627B token cleaned and deduplicated version of RedPajama},'' \url{https://www.cerebras.net/blog/slimpajama-a-627b-token-cleaned-and-deduplicated-version-of-redpajama}, 2023. [Online]. Available: \url{https://huggingface.co/datasets/cerebras/SlimPajama-627B}
\BIBentrySTDinterwordspacing

\bibitem{saparovlanguage}
A.~Saparov and H.~He, ``Language models are greedy reasoners: A systematic formal analysis of chain-of-thought,'' in \emph{The Eleventh International Conference on Learning Representations}, 2023.

\bibitem{arpit2017closer}
D.~Arpit, S.~Jastrzbski, N.~Ballas, D.~Krueger, E.~Bengio, M.~S. Kanwal, T.~Maharaj, A.~Fischer, A.~Courville, Y.~Bengio \emph{et~al.}, ``A closer look at memorization in deep networks,'' in \emph{Proceedings of the 34th International Conference on Machine Learning-Volume 70}, 2017, pp. 233--242.

\bibitem{rahaman2018spectral}
N.~Rahaman, D.~Arpit, A.~Baratin, F.~Draxler, M.~Lin, F.~A. Hamprecht, Y.~Bengio, and A.~Courville, ``On the spectral bias of deep neural networks,'' \emph{International Conference on Machine Learning}, 2019.

\bibitem{xu2019frequency}
Z.-Q.~J. Xu, Y.~Zhang, T.~Luo, Y.~Xiao, and Z.~Ma, ``Frequency principle: Fourier analysis sheds light on deep neural networks,'' \emph{Communications in Computational Physics}, vol.~28, no.~5, pp. 1746--1767, 2020.

\bibitem{xu_training_2019}
Z.-Q.~J. Xu, Y.~Zhang, and Y.~Xiao, ``Training {Behavior} of {Deep} {Neural} {Network} in {Frequency} {Domain},'' in \emph{Neural {Information} {Processing}}, ser. Lecture {Notes} in {Computer} {Science}, 2019, pp. 264--274.

\end{thebibliography}

\section{Biography Section}

\begin{IEEEbiography}[{\includegraphics[width=1in,height=1.25in,clip,keepaspectratio]{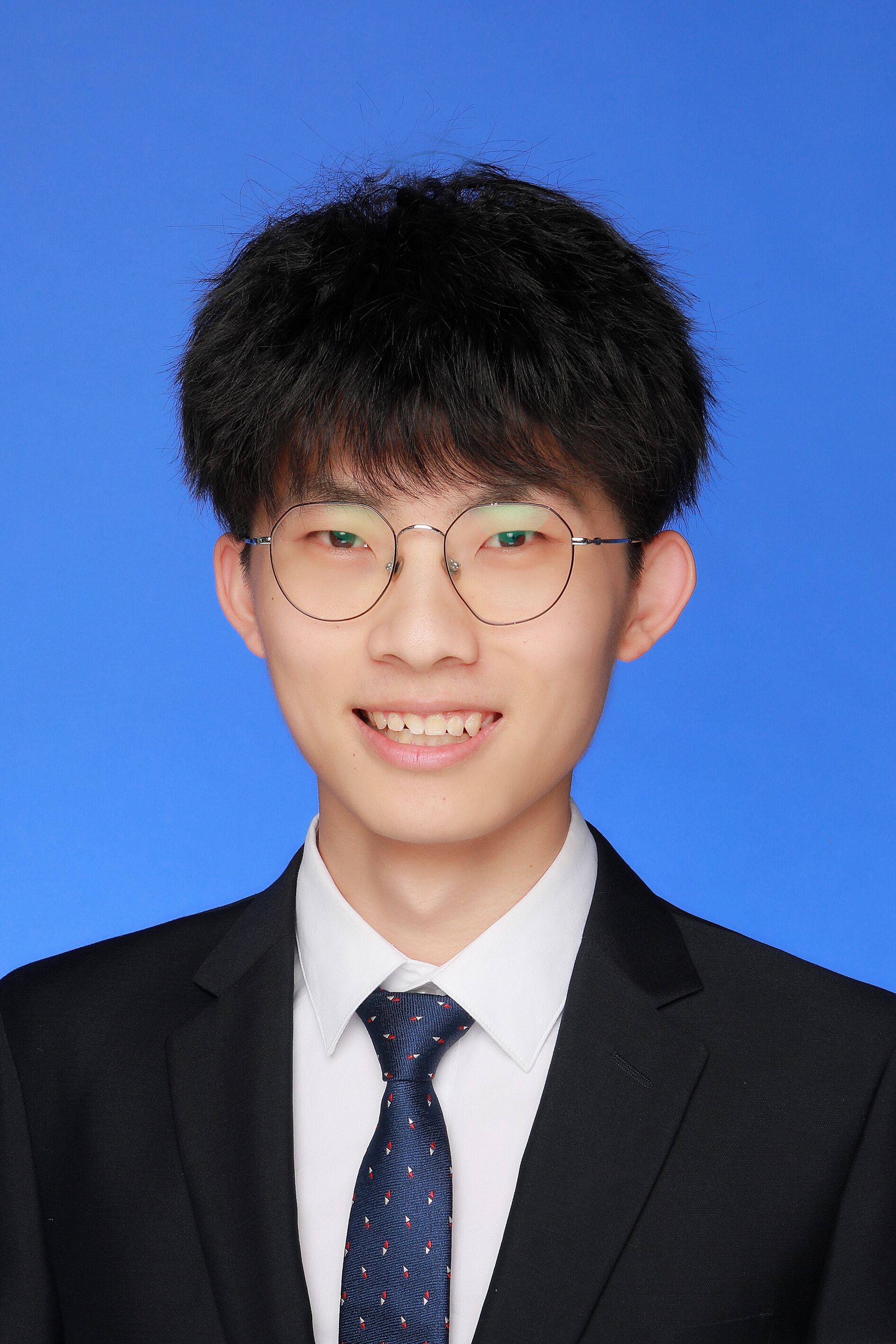}}]{Zhongwang Zhang}

received the bachelor’s degree from Zhiyuan College of Shanghai Jiao Tong University in 2021. He is working toward a doctoral degree at the School of Mathematical Sciences, Shanghai Jiao Tong University. His research interests encompass understanding deep learning through the training process, analyzing loss landscapes, studying generalization, exploring model interpretability, conducting mechanism analysis, and developing various applications.
\end{IEEEbiography}

\begin{IEEEbiography}[{\includegraphics[width=1in,height=1.25in,clip,keepaspectratio]{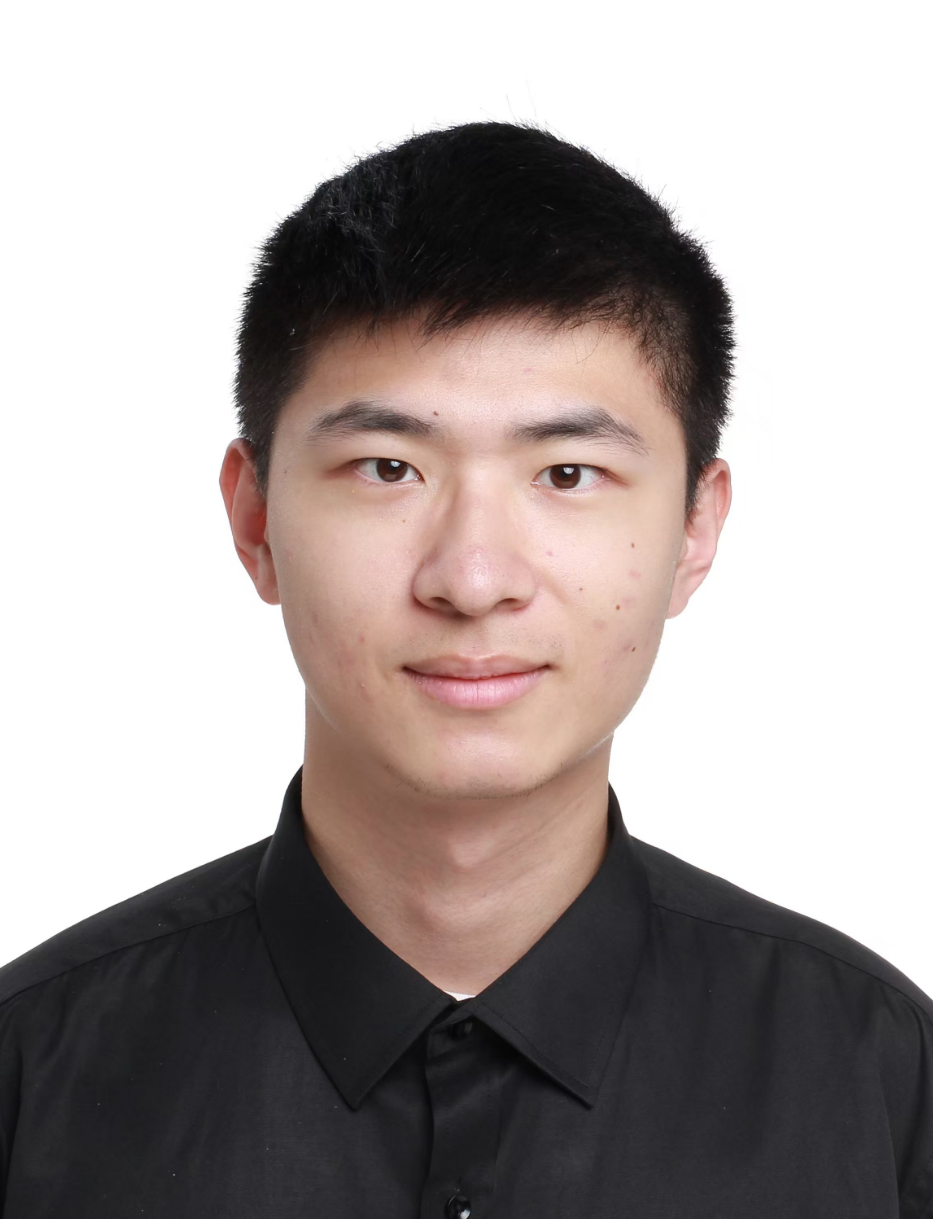}}]{Pengxiao Lin}

received his bachelor's degree in 2023 and is currently pursuing a doctoral degree at the School of Mathematical Sciences, Shanghai Jiao Tong University. His research focuses on natural language processing and language model reasoning. Additionally, he is interested in the techniques and principles behind training large language models.
\end{IEEEbiography}

\begin{IEEEbiography}[{\includegraphics[width=1in,height=1.25in,clip,keepaspectratio]{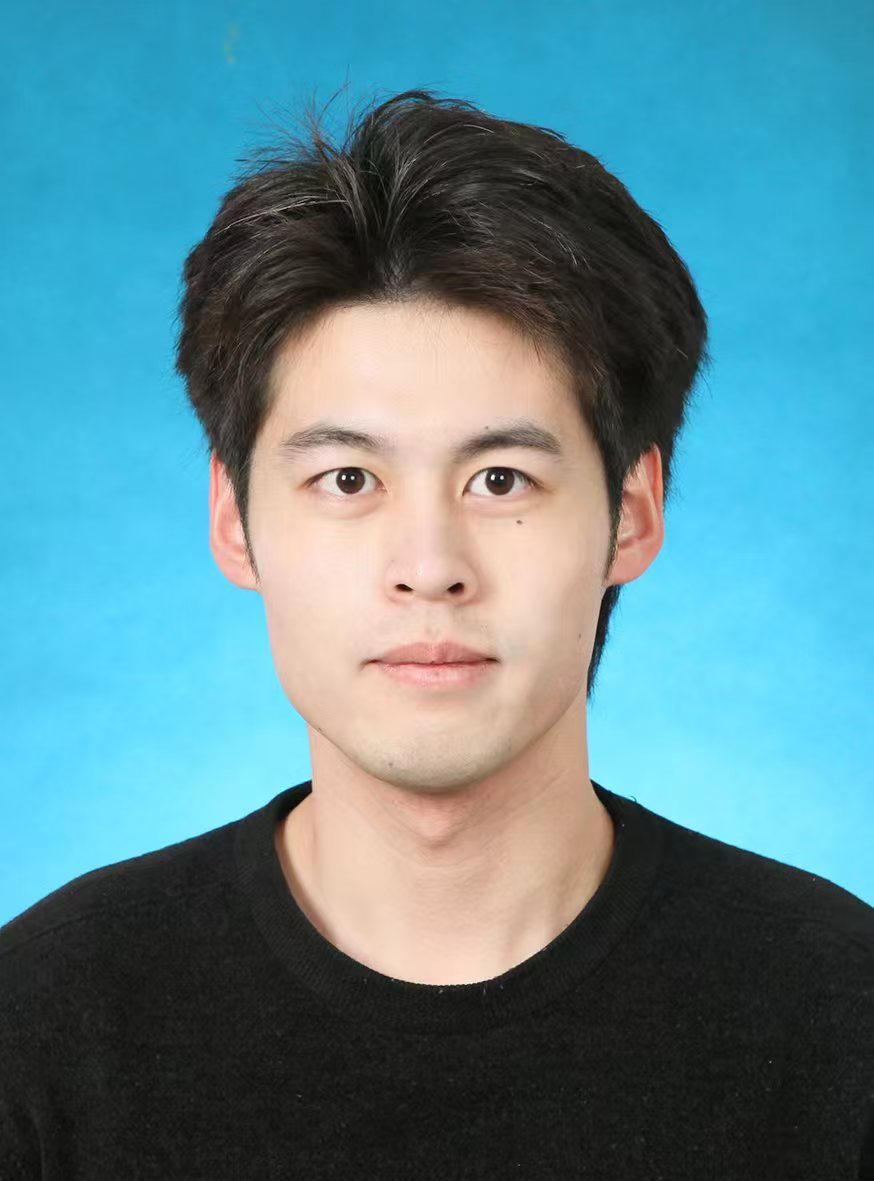}}]{Zhiwei Wang}

received the bachelor’s degree from Zhiyuan College of Shanghai Jiao Tong University in 2021. He is working toward a doctoral degree at the School of Mathematical Sciences, Shanghai Jiao Tong University. His interests include AI for science, understanding deep learning from the training process, mechanism, and generalization.
\end{IEEEbiography}

\begin{IEEEbiography}[{\includegraphics[width=1in,height=1.25in,clip,keepaspectratio]{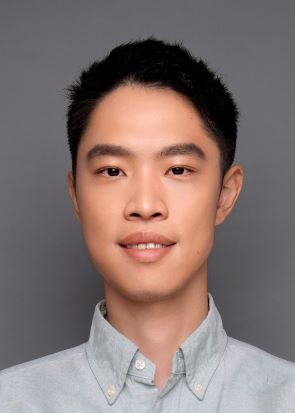}}]{Yaoyu Zhang}
is an associate professor at the Institute of Natural Sciences and the School of Mathematical Sciences, Shanghai Jiao Tong University. He earned his Bachelor's degree in Physics in 2012, and his Ph.D. in Mathematics in 2016 from Shanghai Jiao Tong University. From 2016 to 2020, he conducted postdoctoral research at New York University Abu Dhabi \& Courant Institute, as well as the Institute for Advanced Study in Princeton. His research focuses on the theoretical foundation of deep learning, particularly the nonlinear training dynamics and condensation phenomenon of deep learning.
\end{IEEEbiography}

\begin{IEEEbiography}[{\includegraphics[width=1in,height=1.25in,clip,keepaspectratio]{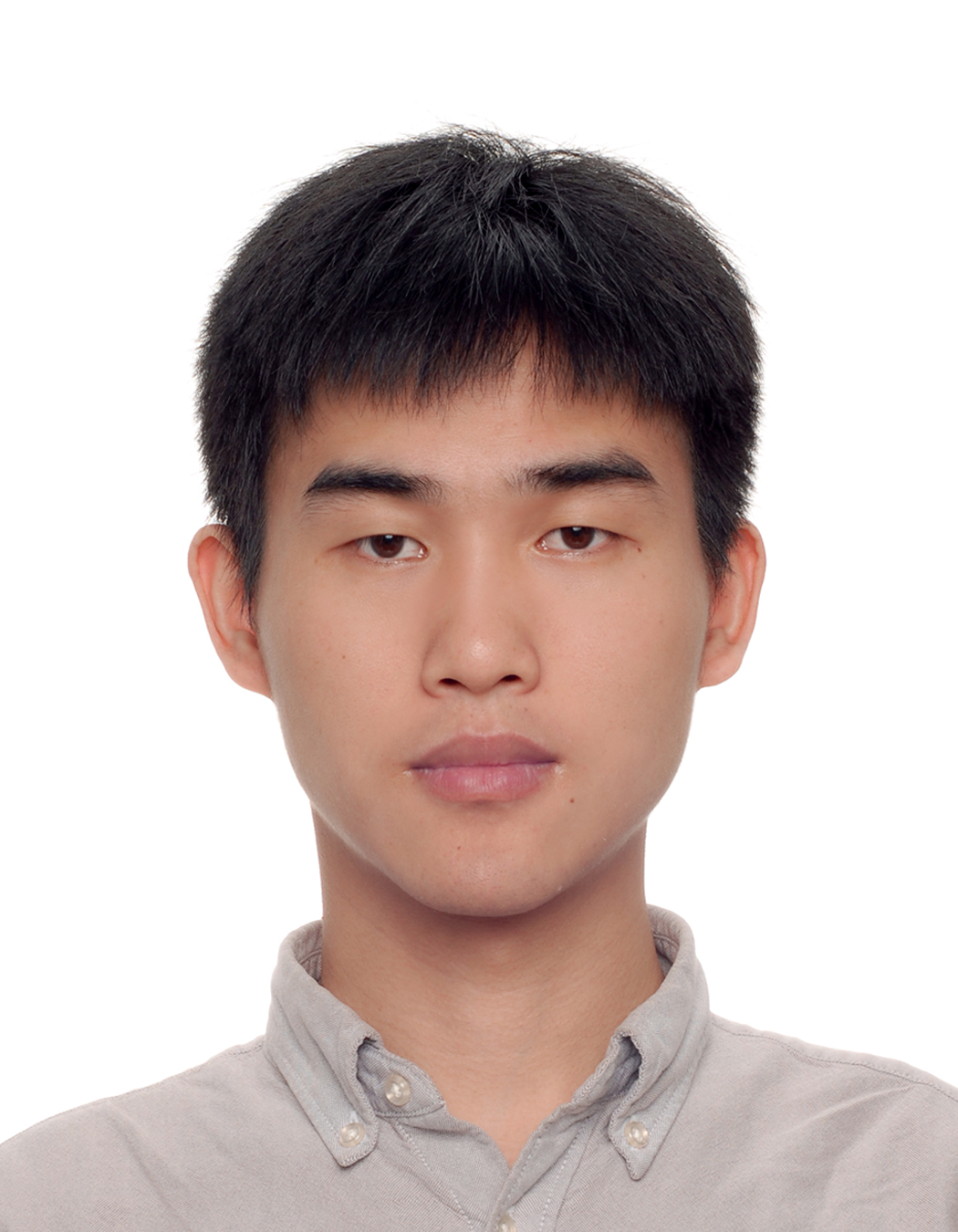}}]{Zhi-Qin John Xu}
is an associate professor at the Institute of Natural Sciences/School of Mathematical Sciences, Shanghai Jiao Tong University. He graduated from Zhiyuan College of Shanghai Jiao Tong University in 2012. In 2016, he graduated from Shanghai Jiao Tong University with a doctor's degree in applied mathematics. From 2016 to 2019, he was a postdoctoral fellow at NYU ABU Dhabi and the Courant Institute. He and collaborators discovered the frequency principle, parameter condensation, and embedding principles in deep learning, and developed multi-scale neural networks. His
interests include understanding deep learning from the training process, loss landscape, generalization, and applications.
\end{IEEEbiography}

\newpage
\appendices

\section{Experimental Setups} \label{app:setup}

For Fig.~\ref{fig:phases}, Fig.~\ref{fig:rank}C, we train the transformer model on a dataset of 900,000 samples, with each input sequence having a length of 9 tokens. The key items in the input sequences are randomly sampled from the range 20-100. We employ the architecture of GPT-2 with 10 layers and 10 heads. The model parameters are initialized using a normal distribution with a mean of 0 and a standard deviation of $(1/d_{\mathrm{in}})^{\gamma}$, where $d_{\mathrm{in}}$ is the input dimension of the parameter and $\gamma$ is the initialization rate. We employ the AdamW optimizer with different weight decay coefficients. The model is trained for 210 epochs using a batch size of 2048 and a gradient clipping maximum norm of 1. The learning rate is scheduled using a warm-up period followed by cosine decay, with a warmup period of 10 epochs, a multiplier of 15 (if there is no special instruction), a cosine decay number of epochs of 200, and a minimum learning rate of 1e-5.

For Fig.~\ref{fig:infor_flow}, Fig.~\ref{fig:rank}A, Fig.~\ref{fig:rank}B, we employ the architecture of a decoder-only transformer with 2 layers and 1 head. Apart from the differences in the model architecture, we retain the same data, training strategy, parameter initialization, and other settings as mentioned in the previous sections. For Fig.~\ref{fig:data_complex}, most of the experimental settings are the same. The difference lies in the dataset size. Specifically, we train the transformer models on a dataset of 100,000 samples. To increase the inferential complexity, we systematically modify the designated target mappings of the following anchor pair groups: \{(a, b), (b, a)\}, \{(a, c), (c, a)\}, \{(a, d), (d, a)\}, \{(b, c), (c, b)\}, \{(b, d), (d, b)\}, and \{(c, d), (d, c)\}. 

For Fig.~\ref{fig:diffu}, we utilized a synthetic dataset comprising 5,000 rendered images of 2D geometric shapes, each annotated with concept classes for size, color, and shape. These images were generated using Blender, featuring single objects on a blank $28\times 28$ background with eight possible attribute combinations. For our model, we employed a conditional diffusion framework based on a U-Net architecture, which incorporates three upsampling and downsampling convolutional layers, and GELU activations. The diffusion process was conditioned on the concept variables, and the model was trained to minimize the mean squared error between the predicted and true Gaussian noise using the Adam optimizer.

For Fig.~\ref{fig:real_task}, we employ the architecture of GPT-2 in our experiments, with the same dataset setups in the original works.

\section{Dataset Splitting Method}\label{data_split}

A straightforward division based on data ranges proves to be impractical. To illustrate, consider a scenario where the range of key tokens in the training set is denoted as $[i, j]$, while in the test dataset, it is represented as $[j+1, k]$. The encoding of data within the interval $[j+1, k]$ is not learned during the neural network training process. As a result, the neural network fails to produce the key token output for the test dataset.

\begin{figure}[ht]
\centering
\includegraphics[width=1.\linewidth]{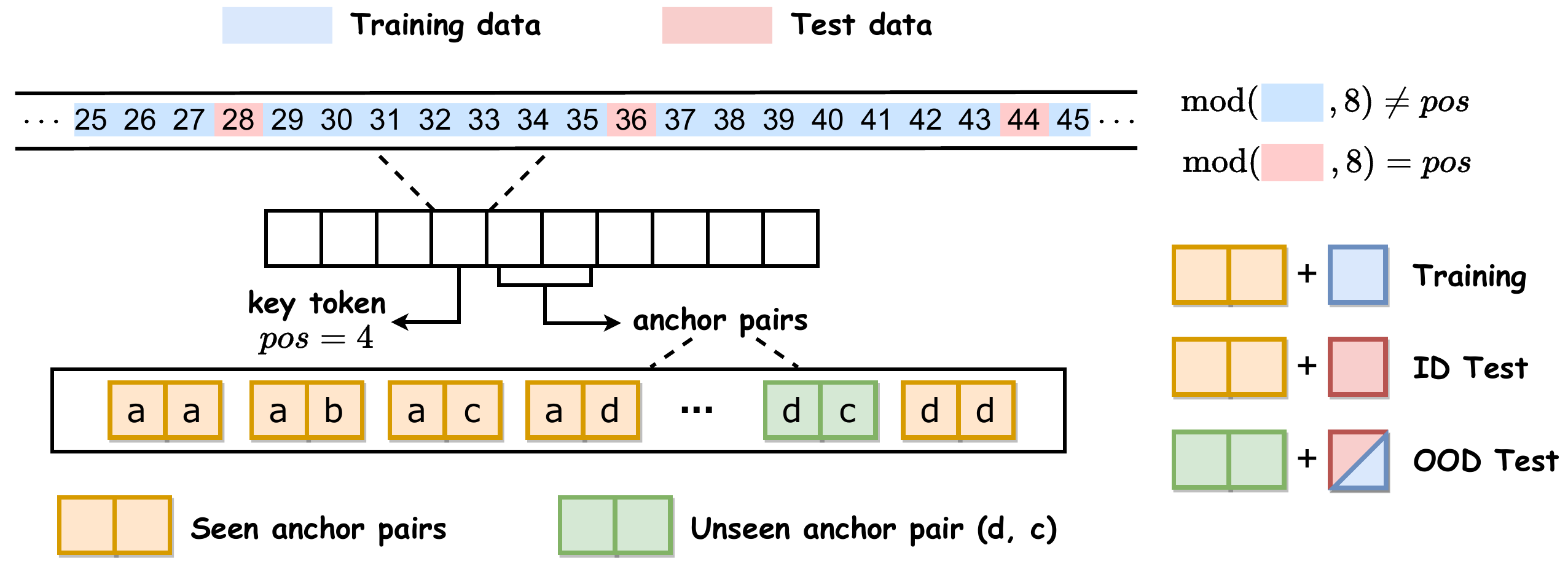}
\caption{Illustration of the dataset splitting method based on the value and position of the key token. The training data and test data are divided according to the modulo operation on the key token value and its position in the input sequence.}
\label{fig:data}
\end{figure}

To address this issue, we divide the dataset based on the value and the position of the key token, as shown in Fig.~\ref{fig:data}. Consider a task with an input sequence of length $n$. For an input sequence in the training dataset, a token $x$ can be placed in the $pos$-th position of such input sequence only when ${\rm mod}(x,n-2)\neq pos$. For an input sequence in the test dataset, if the token at the $i$-th position is a key token, then a token $x$ can be placed in the $pos$-th position of such input sequence only when ${\rm mod}(x,n-2) = pos$. It is important to note that the test data and training data are not completely separated in terms of values. However, when the positions of the key tokens are the same, the corresponding test data and training data do not overlap.

We further define two types of generalization based on this dataset-splitting method. For training data, we use pairs of seen anchors and training data (training key tokens). Regarding data generalization, we test the model using pairs of seen anchors and test data (test key tokens) to evaluate the model's ability to generalize to different tokens within seen composite mappings. For task generalization, we use pairs of unseen anchors with test data or training data to evaluate the model's performance on masked composite mappings. It is important to note that for task generalization, we can test the accuracy of the anchor pairs with different ground truth mappings. This accuracy reflects the model's preferred mappings for these anchor pairs.

\section{Further Experimiental Verification}\label{detailed_condense}

\begin{figure*}[ht]
\centering
\includegraphics[width=0.84\linewidth]{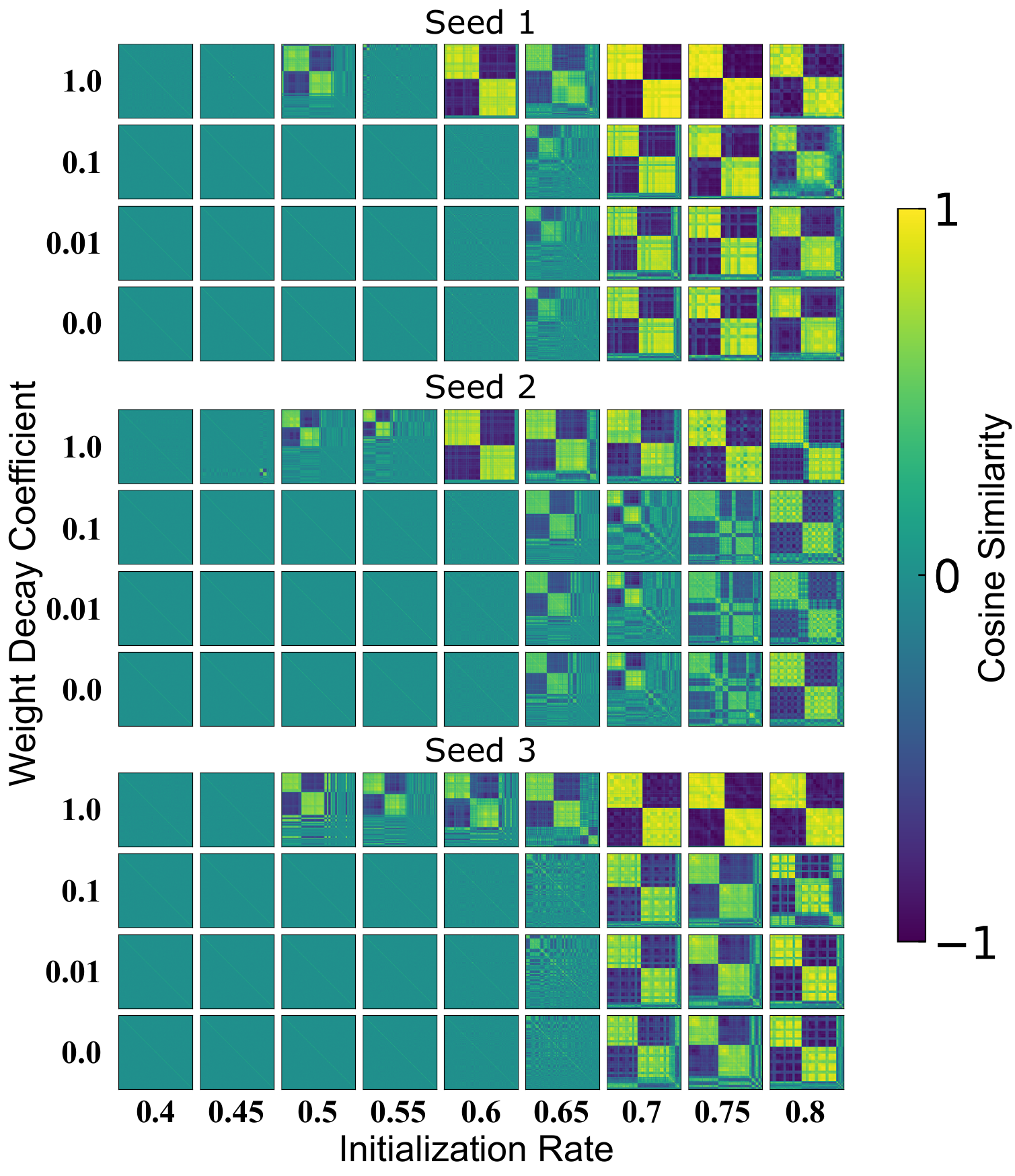}

\caption{Degree of parameter condensation in the parameter matrix \(W^{Q(1)}\) under various settings. The top, middle, and bottom panels correspond to different random seeds. Each subplot depicts the cosine similarity between the input weights of neurons for specific initialization rates and weight decay coefficients. }
\label{fig:condense_all}
\end{figure*}

In this section, we illustrate the degree of parameter condensation in the parameter matrix 
\(W^{Q(1)}\) under various settings. As shown in Fig.~\ref{fig:condense_all}, the top, middle, and bottom panels correspond to different random seeds. Each subplot depicts the cosine similarity between the input weights of neurons for specific initialization rates and weight decay coefficients. It is evident that as the initialization rate and weight decay coefficients increase, the model exhibits greater condensation and reduced parameter complexity.

\vfill

\end{document}